\pdfoutput=1

\documentclass[11pt]{article}

\usepackage[final]{acl}

\usepackage{times}
\usepackage{latexsym}

\usepackage[T1]{fontenc}

\usepackage[utf8]{inputenc}

\usepackage{microtype}

\usepackage{inconsolata}

\usepackage{graphicx}
\usepackage{xcolor}

\usepackage{amsmath}
\usepackage{amssymb}

\usepackage{subcaption}

\usepackage{enumitem}

\usepackage{pifont}
\newcommand{\cmark}{\ding{51}}%
\newcommand{\xmark}{\ding{55}}%

\usepackage{listings}

\lstnewenvironment{code}[1][]%
  {\noindent\minipage{\linewidth}\medskip 
   \lstset{basicstyle=\ttfamily,frame=single,#1}}
  {\endminipage}

\newcommand{\change}[1]{#1}

\title{You Are What You Train: Effects of Data Composition on Training Context-aware Machine Translation Models}

\author{Paweł Mąka \and Yusuf Can Semerci \and Jan Scholtes \and Gerasimos Spanakis\\
        Department of Advanced Computing Sciences \\ 
        Maastricht University \\ 
        \texttt{\{pawel.maka, y.semerci, j.scholtes, jerry.spanakis\}@maastrichtuniversity.nl}}

\begin{document}
\maketitle
\begin{abstract}

Achieving human-level translations requires leveraging context to ensure coherence and handle complex phenomena like pronoun disambiguation. Sparsity of contextually rich examples in the standard training data has been hypothesized as the reason for the difficulty of context utilization. In this work, we systematically validate this claim in both single- and multilingual settings by constructing training datasets with a controlled proportions of contextually relevant examples. We demonstrate a strong association between training data sparsity and model performance confirming sparsity as a key bottleneck. Importantly, we reveal that improvements in one contextual phenomenon do no generalize to others. While we observe some cross-lingual transfer, it is not significantly higher between languages within the same sub-family. Finally, we propose and empirically evaluate two training strategies designed to leverage the available data. These strategies improve context utilization, resulting in accuracy gains of up to 6 and 8 percentage points on the ctxPro evaluation in single- and multilingual settings respectively.\footnote{\url{https://github.com/Pawel-M/data-composition}}

\end{abstract}

\section{Introduction}

Context-Aware Machine Translation (MT) models use surrounding sentences (context) to improve translation by maintaining coherence and resolving ambiguities \citep{agrawal2018contextual, bawden-etal-2018-evaluating, muller-etal-2018-large, voita-etal-2019-good}. The context can be sentences in the source language and the previously translated sentences in the target language. While many works improved the translation quality of the context-aware MT by applying standard Transformer \citep{vaswani2017attention} model \citep{sun-etal-2022-rethinking, majumde2022baseline, gete-etal-2023-works, post-junczys-dowmunt-2024-evaluation, alves2024tower, kocmi-etal-2024-findings}, specialized architectures \citep{tu-etal-2017-context, bawden-etal-2018-evaluating, miculicich-etal-2018-document, maruf-etal-2019-selective, huo-etal-2020-diving, zheng2021towards}, and decoder-only LLMs \citep{alves2024tower, kocmi-etal-2024-findings}, the reason why the context utilization is challenging for the models remain an open question.

\begin{figure}
    \centering
    \includegraphics[width=0.85\linewidth]{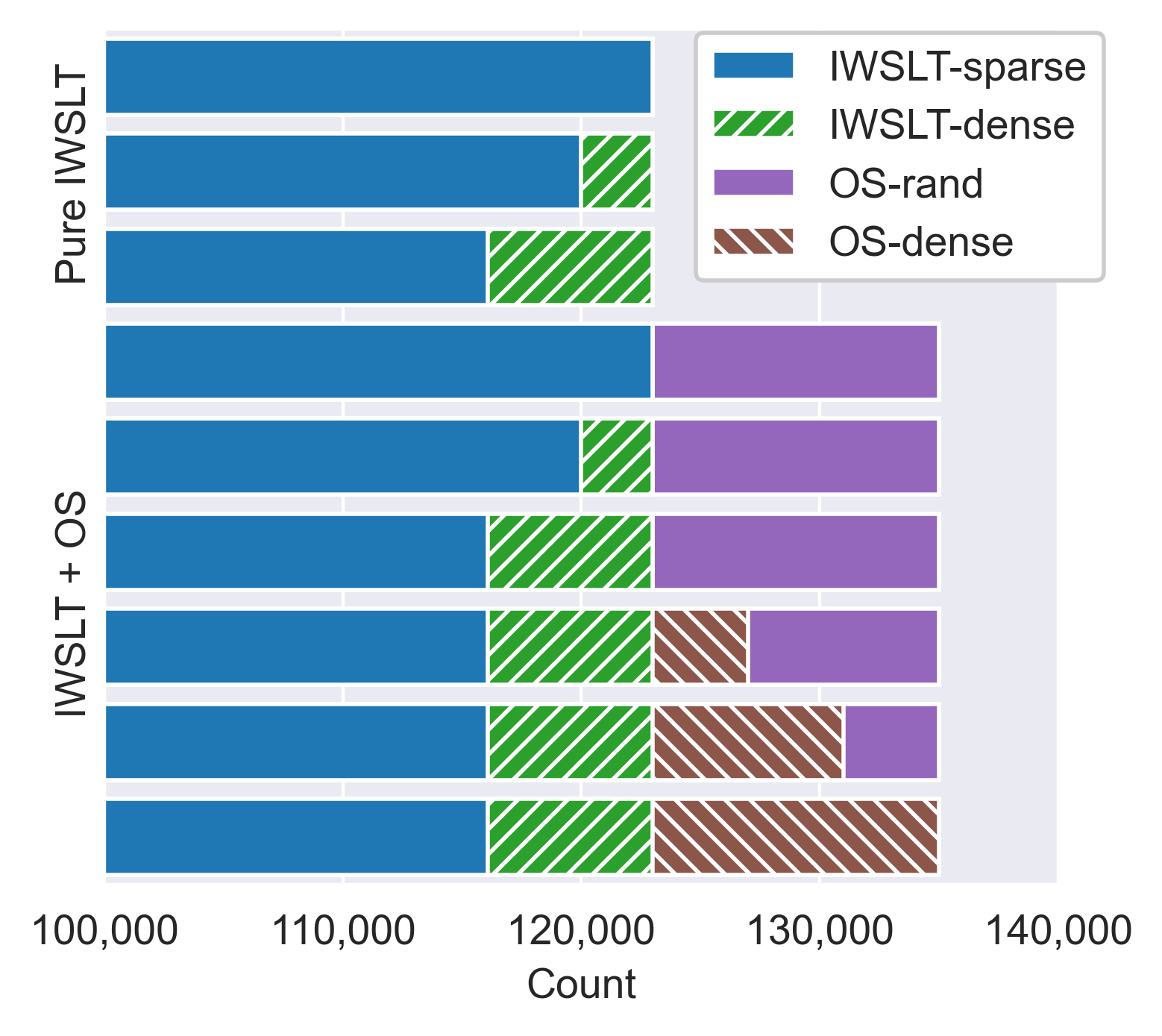}
    \caption{Composition of the English-to-German training datasets with the Gender phenomenon in Pure IWSLT and IWSLT+OpenSubtitles settings. Annotations are based on ctxPro \citep{wicks-post-2023-identifying}, and the dashed bars represent the contextually-rich datasets. Note that the horizontal axis starts at 100,000.}
    \label{fig:dataset-composition-gender}
\end{figure}

The low density of contextually rich (requiring context for correct translation) examples in the training datasets has been suspected as the main reason why MT models have trouble in translating contextual phenomena. For example, \citet{lupo-etal-2022-divide} proposed the two-fold sparsity hypothesis, where the low density of examples in the dataset and the tokens in the examples requiring context increases the difficulty of learning to leverage context. \citet{post-junczys-dowmunt-2024-evaluation} show that sparsity in the evaluation datasets makes it difficult to assess the context utilization of the models. 
We argue that this also points to the sparsity hypothesis in the training data, as the evaluation datasets are often \change{sampled from the same distribution (the underlying dataset)}. 

In this work, we evaluate how the the proportion of contextually rich examples in the training data of the context-aware MT models affects the overall translation quality measured by BLEU \citep{papineni-etal-2002-bleu} and COMET \citep{rei-etal-2020-comet}, and performance on the examples requiring context (using generative and contrastive evaluations). To this end, we use ctxPro toolset \citep{wicks-post-2023-identifying} to extract the relevant examples containing the following phenomena: Gender, Formality, Auxiliary, Inflection, and Animacy. \change{The details of the annotation and phenomena can be found in the original paper \citep{wicks-post-2023-identifying} (see Appendix~\ref{sec:phenomena} for short descriptions).} We constructed training data by mixing contextually rich and poor examples with varying proportions (Figure~\ref{fig:dataset-composition-gender} illustrates this for Gender in English-to-German). Moreover, we evaluate cross-lingual transfer of context utilization in multilingual models on English-to-X and X-to-English where X is \{German, French, Polish, Russian, and Spanish\}. Finally, we explore several ways to effectively leverage the available data to obtain models that perform well both generally and in context-sensitive settings. 
The contributions of this work are:
\begin{enumerate}[topsep=3pt,itemsep=3pt,partopsep=0pt, parsep=0pt, itemindent=15pt, leftmargin=0pt]
    \item We \textbf{empirically validate the sparsity hypothesis}, showing strong relation between the density of the contextual phenomena in the training data and the resulting performance of the context-aware MT models. 
    \item We \textbf{reveal limitations in generalization}, showing that the improvement in one linguistic phenomenon does not transfer to others. We observe limited cross-lingual transfer, not substantially higher between languages in the same sub-family.
    \item We propose and empirically evaluate \textbf{two training strategies} designed to improve context utilization by leveraging the available data.
    We show a trade-off between improving context utilization and general translation metrics such as BLEU.
\end{enumerate}

\section{Related Work}

Through years many dedicated architectures have been proposed for context-aware MT \citep{miculicich-etal-2018-document, voita-etal-2019-good, voita-etal-2019-context, bao-etal-2021-g, chen2022one, feng-etal-2022-learn, bulatov2022recurrent, maka-etal-2024-sequence} including popular multi-encoder (where a separate encoder is responsible for processing the context sentences; \citealp{jean2017does, miculicich-etal-2018-document, maruf-etal-2019-selective, huo-etal-2020-diving, zheng2021towards}), but the standard Transformer model \citep{vaswani2017attention} with the sentences being concatenated (single-encoder; \citealp{tiedemann-scherrer-2017-neural, ma-etal-2020-simple, zhang-etal-2020-long}) exhibited high performance despite its relative simplicity \citep{majumde2022baseline, sun-etal-2022-rethinking, gete-etal-2023-works, post2023escaping}. While decoder-only LLMs have achieved state-of-the-art results in MT \citep{alves2024tower, kocmi-etal-2024-findings}, they require extensive datasets for training, have a large number of parameters, and increased inference time \citep{pang-etal-2025-salute}, which can limit their usefulness in computationally constrained environments. In recent years, research interest in the architectures other than decoder-only has remained relevant \citep{mohammed-niculae-2024-measuring, warner2024smarter, alastruey-etal-2024-unveiling, azeemi-etal-2025-label, marashian-etal-2025-priest}. Therefore, we largely focus this paper on encoder-decoder models.

The standard sentence-level metrics (e.g., BLEU \citep{papineni-etal-2002-bleu} do not capture the contextual utilization by the models \citep{hardmeier2012discourse, wong-kit-2012-extending}. To address this, several evaluation datasets have been proposed including contrastive \citep{muller-etal-2018-large, bawden-etal-2018-evaluating, voita-etal-2019-good, lopes-etal-2020-document} and generative such as ctxPro \citep{wicks-post-2023-identifying} used in this study. Moreover, metrics like CXMI \citep{fernandes-etal-2021-measuring} and PCXMI \citep{fernandes-etal-2023-translation} can measure how much the model relies on context during translation.

The effects of the training dataset on the final model has also been studied extensively \citep{kaplan2020scaling, hoffman2022training} in different domains \citep{alabdulmohsin2023getting}, including document-level MT \citep{zhuocheng-etal-2023-scaling}. The studies mostly concentrated on the scale of the training dataset. We, instead, investigate the composition of the dataset and its effect on the context-aware MT models.

Several works proposed methods increasing contextual capabilities of the models by training the models on annotated data \citep{jwalapuram-etal-2020-pronoun, yin-etal-2021-context, gete-etal-2023-targeted, maka-etal-2025-analyzing} but they target only pronoun disambiguation. Fine-tuning in this case can be seen as similar to domain adaptation \citep{luong-manning-2015-stanford, chu-etal-2017-empirical} where loss weighting (similar to one of our methods) is an effective strategy \citep{wang-etal-2017-instance}. 


\section{Effects of Data Composition}

We first measured how the presence of contextually rich examples in the training data affects both translation quality and the models’ ability to leverage context. To that end, we trained models on datasets whose composition we systematically varied. Specifically, we identified contextual examples (containing relevant phenomena) from the available datasets using ctxPro toolset \citep{wicks-post-2023-identifying} and constructed a series of datasets with varying densities of different phenomena. This setup allowed us to assess inter-phenomena as well as cross-lingual effects of the composition of the training datasets. \change{We used three settings: single language pair (English-to-German), and multilingual with encoder-decoder and decoder-only (LLM) models.} For the multilingual setting, we used English-to-X and X-to-English  language directions, where X is \{German, French, Polish, Russian, and Spanish\} - a subset of directions covered by the ctxPro. We utilized two Germanic, Romance, and Slavic languages.

\subsection{Datasets}

We base our research on two document-level translation datasets: IWSLT 2017 English-to-German \citep{cettolo-etal-2017-overview} and OpenSubtitles 2018 \citep{lison-etal-2018-opensubtitles2018}. For the English-to-German direction, we employ both datasets, and for the multilingual setting, we only use OpenSubtitles. We extract contextual annotations from the training subset of the IWSLT dataset using the ctxPro toolset. The annotated (containing contextually-rich examples) subset forms \textbf{IWSLT-dense} dataset, which can be further divided based on the target phenomenon: Gender, Formality, Auxiliary, Inflection, and Animacy. \change{We discard examples containing more than one type of phenomena in any of the sentences.} From the remaining examples we form \textbf{IWSLT-sparse} dataset of size 123,000, containing no examples annotated with \change{any} contextual phenomena. CtxPro released annotations extracted from the OpenSubtitles 2018 dataset divided into \textit{dev}, \textit{devtest}, and \textit{test} subsets. We set aside the \textit{test} subset for the evaluation and used the combined \textit{dev} and \textit{devtest} subsets for training, forming \textbf{OS-dense} dataset. The released ctxPro dataset is not exhaustive; therefore, we do not create the sparse version of the OpenSubtitles dataset. Instead, we randomly sample the OpenSubtitles dataset to the desired size (referred to as \textbf{OS-random}). \change{It should be noted that OS-random datasets can contain a very limited number of examples from OS-dense datasets (less than 1 per 1000).}
In Appendix~\ref{sec:datasets-composition} we present the sizes of the dense component datasets.

\begin{figure*}
    \centering
    \includegraphics[width=1\linewidth]{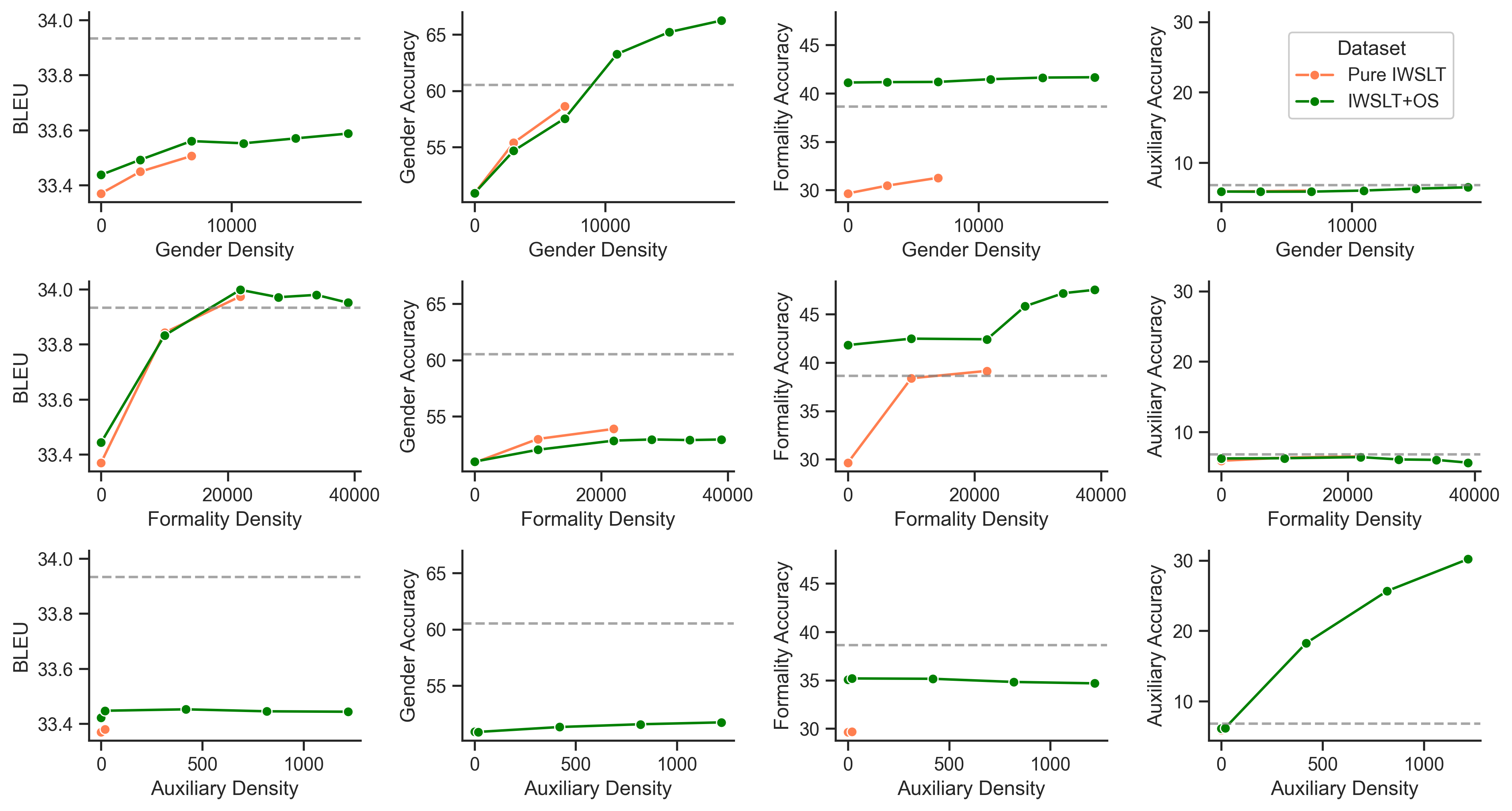}
    \caption{Measured metrics of BLEU on IWSLT 2017 testset, and ctxPro accuracy on Gender, Formality, and Auxiliary phenomena (in columns) of the \change{OpusMT en-de} models trained on the datasets with varying amounts of contextually-rich examples of Gender, Formality, and Auxiliary phenomena (in rows). Shows two experimental settings: Pure IWSLT and combined IWSLT+OS.}
    \label{fig:sparsity-opusmt-results}
\end{figure*}

To create the training datasets with varying densities of contextually rich examples, we sample and concatenate examples from both dense and sparse datasets to form a training dataset. For English-to-German, we study two settings: \textit{Pure IWSLT} (only IWSLT-sparse and IWSLT-dense datasets) and \textit{IWSLT + OS} (using IWSLT-sparse, IWSLT-dense, and English-to-German OS-rand and OS-dense datasets). These allow us to study two regimes: extremely low sparsity with the first setting, and very dense with the second one. We progressively replace examples from sparse and random datasets with the examples sampled from dense datasets. In the multilingual experiments, we formed the baseline training dataset by sampling 50,000 examples from OS-rand for all language directions we considered. For each phenomenon in a language direction, we formed the enriched datasets by replacing $n$ examples with the examples sampled from the OS-dense dataset corresponding to the phenomenon and language direction. We chose $n$ to be the minimum number of examples (rounded) for a particular phenomenon across language directions maximizing the resulting density of the training datasets while making the results comparable between language directions. We present the illustration of the composition of the datasets in Figure~\ref{fig:dataset-composition-gender} for Gender on English-to-German and further details in Appendix~\ref{sec:datasets-composition}. \change{To reduce the complexity of the analysis we add only examples containing a single type of phenomenon. Assessing the complex interconnections between phenomena is left for future work.}

\subsection{Training}
\change{For encoder-decoder models,} we employed a two-stage training process where first the sentence-level model is trained on more abundant sentence-aligned datasets, followed by the context-aware training on the document-level dataset. Following \citet{maka-etal-2025-analyzing}, we rely on the publicly available pre-trained sentence-level models, namely \textit{OPUS-MT en-de}
\citep{TiedemannThottingal:EAMT2020, tiedemann2023democratizing}
and \textit{No Language Left Behind} (NLLB-200) with 600M parameters
\citep{nllb2022}. \change{For LLM-based MT models, we utilize Towerbase 7B model \cite{alves2024tower} which we fine-tune using LoRA \cite{hu2022lora} on document-level MT dataset. Because Towerbase models were not pre-trained on Polish language we do not include English-to-Polish and Polish-to-English language pairs in training and evaluation.}
We concatenate consecutive sentences separated by the special \verb|[SEP]| token \change{in case of encoder-decoder models and} \verb|<sep>| \change{string in case of LLMs} on both the source and target sides. Similar to \citet{sun-etal-2022-rethinking}, we create examples with all context sizes (number of previous sentences to concatenate) from zero to the maximum context size. 
We set the maximum context size to three as further increases have shown diminishing returns regarding context utilization \citep{post2023escaping}. \change{In Appendix~\ref{sec:phenomena}, we show the number of examples in the ctxPro dataset with antecedent distance inside the context size.} During inference, the models receive only the source-side context and generate the target-side context before the current sentence. We obtained the translation of the current sentence by splitting the output on the separator token \change{(for encoder-decoder models) or substring (for LLMs)}. The training hyper-parameters and additional details can be seen in Appendix~\ref{sec:training-hyperparameters}.

\subsection{Single Language Pair Results}

For the models in the English-to-German experiments, we trained 5 models with different seeds and averaged the results. Apart from the constructed datasets, we also trained a baseline model on the unmodified IWSLT training dataset. To measure the general translation quality, we translated the IWSLT 2017 English-to-German test subset (with BEAM search of 5) and measured BLEU \citep{papineni-etal-2002-bleu}. Additionally, we translated test subsets of the ctxPro dataset (based on OpenSubtitles) and measured the accuracy of matching the expected word in the translation (using the scripts provided with the dataset). The results can be seen in Figure~\ref{fig:sparsity-opusmt-results}. Extended results including COMET and ContraPro \citep{muller-etal-2018-large} accuracy can be found in Appendix~\ref{sec:extended-sparsity-results}.

\begin{figure*}
    \centering
    \includegraphics[width=1\linewidth]{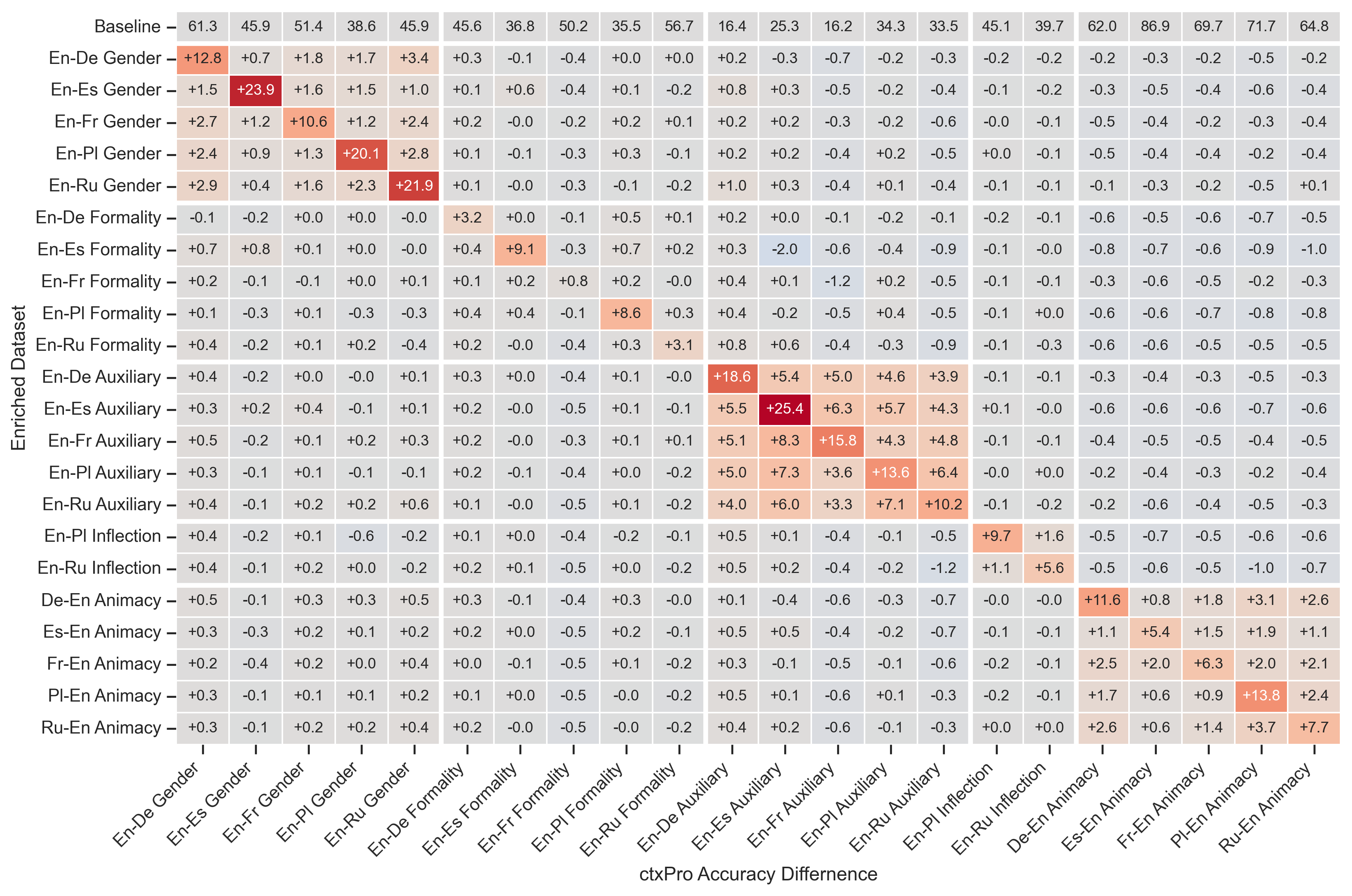}
    \caption{Accuracy on all phenomena for each relevant language direction in ctxPro (in columns) of the \change{NLLB-200 600M models} trained on the OpenSubtitles datasets with varying amounts of contextually-rich examples for each phenomenon and language direction (in rows). We show the differences from the baseline model (top row).}
    \label{fig:sparsity-nllb-results}
\end{figure*}

We observed a drop in BLEU for the models trained on the sparse datasets, even for the datasets with mixed OpenSubtitles examples. While the reduction was relatively small (less than 2\%), it returned to the baseline value only when Formality IWSLT-dense examples were added to the dataset. This could mean that the examples from the IWSLT dataset annotated with Formality were particularly influential for the model's general translation ability, and mixing in the random examples from OpenSubtitles did not help.

For Gender and Formality, increasing their density in the training dataset improved the ctxPro accuracy for the corresponding phenomenon. Notably, Formality in the IWSLT+OS setting only improved when OS-dense examples were added, but exceeded the accuracy of the baseline model even with the most sparse dataset. Adding OS-dense examples improved the accuracy significantly above the baseline (up to 30\%). Interestingly, adding dense examples in one phenomenon had minimal effect on the accuracy of the other phenomena, with only a very small increase of Formality for the Gender-enriched dataset and vice versa. Those results show that the generalizability of the models' ability to handle contextual phenomena is very limited.
\change{While we argue that experimenting with the publicly-available pre-trained models enhances reproducibility OpusMT was trained on OpenSubtitles dataset on which ctxPro dataset was based. Therefore, we include the results where the weights has been randomly initialized in Appendix~\ref{sec:extended-sparsity-results} which show the same behavior corroborating our findings.}

\begin{figure*}
    \centering
    \includegraphics[width=0.83\linewidth]{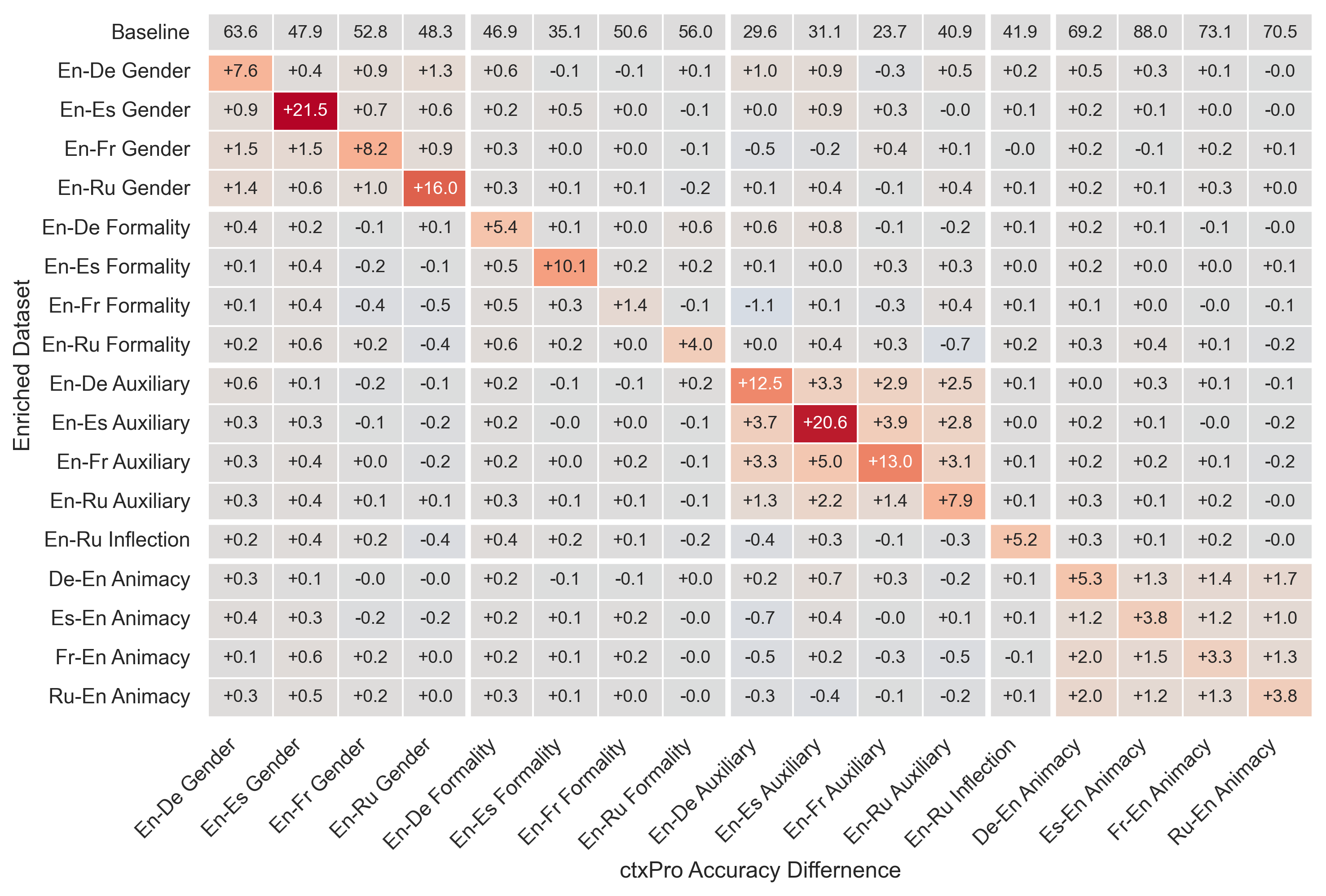}
    \caption{\change{Accuracy on all phenomena for each relevant language direction in ctxPro (in columns) of the Towerbase 7B trained on the OpenSubtitles datasets with varying amounts of contextually-rich examples for each phenomenon and language direction (in rows). We show the differences from the baseline model (top row).}}
    \label{fig:sparsity-towerbase-results}
\end{figure*}
\subsection{Multilingual Results}

For the multilingual experiments, we trained models (with a single seed due to the computational cost of training and evaluation) on the composed datasets and measured ctxPro accuracy for all applicable phenomena and language directions included in the experiments. Note that Inflection applies only to English-to-Polish and English-to-Russian, and Animacy only to X-to-English. \change{The results are presented in Figures~\ref{fig:sparsity-nllb-results} and ~\ref{fig:sparsity-towerbase-results} for encoder-decoder and decoder-only models respectively.} Results in terms of BLEU and COMET on the testsets sampled from OpenSubtitles for each language direction can be seen in Appendix~\ref{sec:extended-sparsity-results}.

For each model, the highest improvement in accuracy was observed for the phenomenon and language direction that was added to the training dataset (values on the diagonal in the figures). In line with the results on the single language pair, we did not observe any intra-lingual transfer between phenomena. Interestingly, there was some transfer between language directions for the same phenomenon, which was the strongest for Auxiliary, moderate for Gender, Inflection \change{(for encoder-decoder models)}, and Animacy, and no transfer for Formality. Contrary to our expectations, we did not observe notably stronger transferability between languages in the same linguistic sub-family, with the exception of Auxiliary in encoder-decoder models, where the increase in accuracy is slightly higher inside Romance and Slavic languages than for other languages. \change{Surprisingly, Towerbase did not exhibit higher generalizaility compared to NLLB-200 corroborating the notion that LLMs are a reflection of their training data.}

\subsection{Discussion}

We experimentally confirmed the dataset sparsity hypothesis by showing that the models trained on datasets sparse in contextually rich examples exhibit poor context utilization, and increasing the density leads to large improvements for the tested phenomena.
Our experiments showed that the models do not generalize context utilization between phenomena. This finding calls for caution when interpreting the results of evaluations targeting a single phenomenon \citep{muller-etal-2018-large, lopes-etal-2020-document}. While document-level training datasets typically include a representative (for a particular domain) mixture of contextual phenomena, we found that models can develop strong capabilities for some phenomena, while remaining weak on others. \citet{maka-etal-2025-analyzing} found attention heads in context-aware MT models responsible for pronoun disambiguation with some cross-lingual behavior, which is in line with the observed transferability between language directions. We hypothesize that the poor transfer between phenomena can be explained by the models developing separate heads for each of them.

\section{Methods Exploiting Contextual Data}

Inspired by the fact that increased density in contextually-relevant examples of the training dataset leads to improvement in context utilization, we tested several techniques that could leverage the available data more efficiently. We broadly divide them into annotation-based and annotation-free. Annotations can inform the training process but require an external tool (e.g., ctxPro) to mark the relevant examples. A straightforward method is to simply extract the annotated examples from the training dataset and use them to fine-tune the model. Annotation-free methods do not rely on an external tool and have the advantage of generalizability beyond the phenomena covered by any tool. Crucially, the presented methods aim to improve contextual capabilities without the need for any additional data beyond the standard training datasets.

\subsection{Token-level Loss Weighting}
We adapted the weighting of the loss elements \cite{wang-etal-2017-instance}, which increases the error signal coming from selected examples. Instead of weighting the whole examples, we apply a token-level approach as phenomena annotations contain an expected word or phrase that requires context for successful translation. We train the models using the weighted negative log-likelihood loss function:
\begin{equation} \label{eq:weighted-loss}
    \mathcal{L} = - \frac{1}{|D_a|} \sum_{(x_i, y_i, a_i) \sim D_a} \sum_{j=1}^{|y_i|} w(a_{i, j}) log(\hat{y}_{i,j}),
\end{equation}
where $\hat{y}_{i,j}$ is the probability of the $j$-th token in $i$-th example, $D_a$ is the annotated training dataset with examples containing input and output sequences ($x_i$ and $y_i$ respectively), as well as the token-level annotations $a_i$ marking the contextually-dependent tokens, and $w(a_{i,j})$ is defined as:
\begin{equation} \label{eq:weight}
w(a_{i,j}) = \begin{cases}
1 + \lambda, &\text{if contextually dependent}, \\
1, &\text{otherwise}, \\
\end{cases}
\end{equation}
for each token $j$ in the $i$-th output sequence, where $\lambda$ is the hyper-parameter.

\subsection{Metric-based Example Selection}

A major issue with using annotations is that, according to our experiments on data composition, the model will improve only on the included phenomena. To mitigate this, we propose to utilize the model itself to mark contextually-rich examples. \citet{fernandes-etal-2023-translation} proposed the Point-wise Cross-Mutual Information (PCXMI) metric to measure the context reliance of the translations, which is based on the output probabilities of the context-aware MT model. For a particular example it is calculated as:
\begin{equation}\label{eq:pcxmi}
    PCXMI = \sum_{j=1}^{|y|} log \frac{q(y_{j}|y_{t<j}, x, C)}{q(y_{j}|y_{t<j}, x)},
\end{equation}
where $C$ is the context, and $q$ represents the context-aware MT model (returning token probabilities, noted as $q(y_{j}|y_{t<j}, x, C)$) that is trained to also be used as a sentence-level model (noted as $q(y_{j}|y_{t<j}, x)$). We introduce a slightly modified metric that computes the maximum token-level PCXMI for a given example:
\begin{equation}\label{eq:max-pcxmi}
    MaxPCXMI = \max_{j} \Bigl( log \frac{q(y_{j}|y_{t<j}, x, C)}{q(y_{j}|y_{t<j}, x)} \Bigr).
\end{equation}
We motivate it by the fact that an example with even a single token being dependent on context can be considered a contextually-rich example (certainly the case for pronouns), which is better captured by our metric. 
The proposed method consists of the following steps: 
\begin{enumerate}[topsep=3pt,itemsep=3pt,partopsep=0pt, parsep=0pt, itemindent=15pt, leftmargin=0pt]
    \item \textbf{train} the model on context-aware data,
    \item \textbf{calculate} the metric using the trained model for the examples in the training dataset,
    \item \textbf{select} top $k$ examples (a hyper-parameter),
    \item \textbf{fine-tune} the model on the selected subset.
\end{enumerate}

While the method can be seen as similar to curriculum learning \citep{zhang2018empirical}, we select the examples that the model is already competent at translating using context. Intuitively, this is a positive feedback where the model learns to generalize to the difficult examples by becoming better at what it already knows.

\begin{figure*}
    \centering
    \includegraphics[width=1\linewidth]{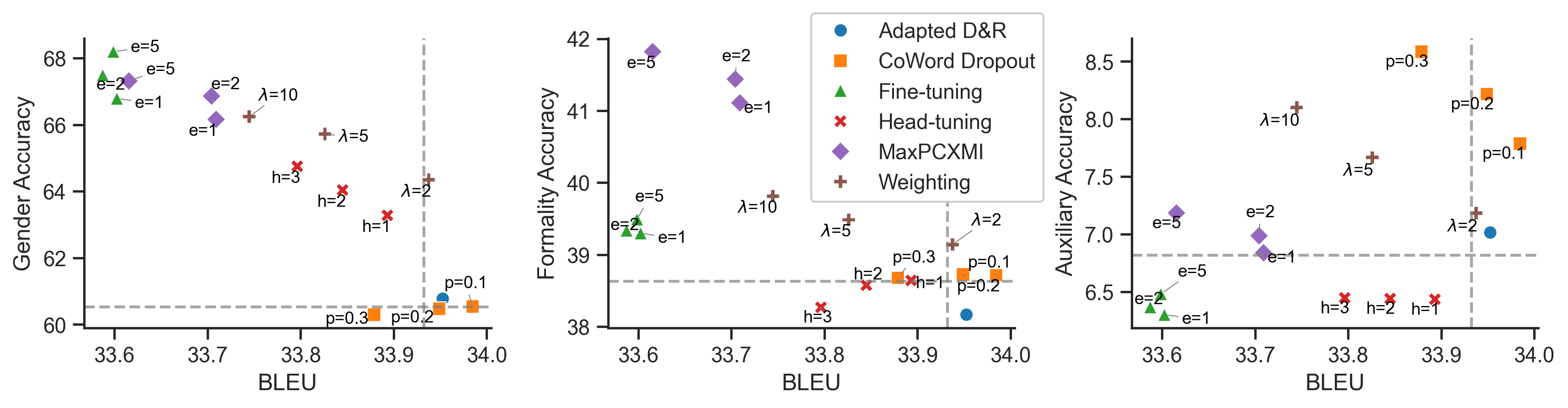}
    \caption{Accuracy of ctxPro English-to-German phenomena against BLEU on the IWSLT 2017 en-de testset of the methods \change{applied to OpusMT en-de model}. Labels show: the number of epochs ("e"), CoWord Dropout probability ("p"), number of tuned heads ("h"), and weighting strength ("$\lambda$") hyper-parameters.}
    \label{fig:finetuning-opusmt}
\end{figure*}

\section{Experiments}

\begin{table}
    \centering
    \begin{tabular}{lcc}
        \hline
        \textbf{Method} & \textbf{Requires} & \textbf{Additional} \\
         & \textbf{Annotations} & \textbf{Training} \\
        \hline
        Fine-tuning & \cmark & \cmark \\
        Adapted D\&R & \xmark & \xmark \\
        CoWord Dropout & \xmark & \xmark \\
        Head-tuning & \cmark & \cmark \\
        \hline
        Weighting & \cmark & \xmark \\
        MaxPCXMI & \xmark & \cmark \\
        \hline
    \end{tabular}
    \caption{\change{Tested methods and whether they require annotated dataset or employ additional fine-tuning.}}
    \label{tab:methods-details}
\end{table}

We experimentally evaluated Token-level Loss Weighting and Metric-based Example Selection for fine-tuning on encoder-decoder models and compared them to the following baselines \change{(Table~\ref{tab:methods-details} summarizes their requirement of annotated dataset and additional training)}:
\begin{itemize}[topsep=3pt,itemsep=3pt,partopsep=0pt, parsep=0pt, itemindent=15pt, leftmargin=0pt]
    \item \textbf{Fine-tuning} (annotation-based) - simply fine-tuning the model on the annotated data after the context-aware training.
    \item \textbf{CoWord Dropout} (annotation-free; \citealp{fernandes-etal-2021-measuring}) - masking random tokens in the current source sentence to force the model to use context for translation, the probability of masking a token is controlled by the hyper-parameter $p$.
    \item \textbf{Adapted Divide and Rule} (annotation-free; \citealp{lupo-etal-2022-divide}) - splitting the current source and target sentences in the middle and appending the first parts to the context. Notably, this method was introduced for the multi-encoder architecture \change{where a separate encoder was used for context sentences. Contextual parameters were trained only in the second, context-aware phase of training with the rest of the model frozen. We \textit{adapt} it to the single-encoder architectures we use in this study by training the whole model in the context-aware training phase}.
    \item \textbf{Head-tuning} (annotation-based; \citealp{maka-etal-2025-analyzing}) - training selected attention heads to attend the context cue, available only for Gender.
\end{itemize}

We evaluated all methods in the single language pair (English-to-German) setting and annotation-free methods in the encoder-decoder multilingual setting (due to the lack of exhaustive annotations for the dataset; see Table~\ref{tab:methods-details}). We used the same base sentence-level models: OpusMT en-de and NLLB-200 600M, respectively. For English-to-German, we trained on the full IWSLT 2017 en-de dataset with ctxPro annotations, and for multilingual, we sampled 50,000 examples for each language direction from the OS-rand dataset. We used the same hyper-parameters shared by all methods as in previous experiments (see Appendix~\ref{sec:training-hyperparameters} for more details) for both training and fine-tuning with the exception of Head-tuning where we applied the hyper-parameters from the original paper. 
In the English-to-German setting, we repeated the training 5 times with different seeds and averaged the results. In the multi-lingual setting, we performed a single training run for all encoder-decoder models with the same seed. 
Fine-tuning used the base model trained with the corresponding seed.

\subsection{Single Language Pair Results}

We tested several parameters for most methods. For fine-tuning-based models, we trained for $e \in \{1, 2, 5\}$ epochs and utilized only the examples with the maximum context size. For Weighting we set the $\lambda$ parameter to 2, 5, and 10. In addition to the values of $p$ for CoWord Dropout recommended by the authors (0.1, 0,2), we also included the value of 0.3. For Metric-based example selection, we set k=30,000 based on the number of annotated examples in the dataset, and used the MaxPCXMI metric (in Appendix~\ref{sec:extended-fine-tuning-results} we present the comparison to the PCXMI metric). For Head-tuning we selected top $h \in \{1, 2, 3\}$ heads from \citet{maka-etal-2025-analyzing}. Results in terms of accuracy on the ctxPro dataset and BLEU on the IWSLT testset can be seen in Figure~\ref{fig:finetuning-opusmt}. Extended results are presented in Appendix~\ref{sec:extended-fine-tuning-results} \change{and calculations of statistical significance of the results can be seen in Appendix~\ref{sec:signifficance}}.

\begin{table*}[ht]
    \centering
    \begin{tabular}{lrrrrrr}
        \hline
        \textbf{Model} & \textbf{BLEU} & \textbf{Gender} & \textbf{Formality} & \textbf{Auxiliary} & \textbf{Inflection} & \textbf{Animacy} \\
        \hline
        Adapted D\&R & \textbf{-0.05} & -0.06 & -0.19 & -0.16 & -0.03 & +0.07 \\
        CoWord p=0.1 & -0.09 & +0.02 & -0.16 & +0.35 & -0.10 & +0.16 \\
        CoWord p=0.2 & -0.11 & +0.07 & -0.28 & +0.65 & -0.21 & +0.01 \\
        CoWord p=0.3 & -0.08 & +0.01 & -0.42 & +0.97 & -0.29 & -0.27 \\
        \hline
        MaxPCXMI e=1 & -0.42 & +1.13 & +0.05 & +3.41 & +0.44 & +1.08 \\
        MaxPCXMI e=2 & -0.45 & +1.42 & +0.05 & +4.25 & +0.57 & +1.10 \\
        MaxPCXMI e=5 & -0.50 & \textbf{+1.93} & \textbf{+0.11} & \textbf{+5.80} & \textbf{+0.76} & \textbf{+1.64} \\
        \hline
    \end{tabular}
    \caption{The averaged (over language directions) difference from the baseline in terms of BLEU on OpenSubtitles 2018 testsets and ctxPro phenomena accuracies for the tested methods \change{applied to NLLB-200 600M model}. Number of epochs is noted as "e", and CoWord Dropout probability  as "p".}
    \label{tab:nllb-fine-tuning-results}
\end{table*}

It can be seen that with four metrics, the models' performance varies, and improvement in one metric comes at a cost of a reduction in another. In particular, we observe a negative relation between ctxPro accuracies and BLEU for all methods with the increase of the hyper-parameters. \change{This necessitates examining the Pareto front in order to assess the performance of the methods.} Metric-based example selection achieved highest improvement in Formality and outperformed the annotation-based selection for fine-tuning in Formality and Auxiliary, and achieved similar results for Gender, with a smaller decrease in BLEU. Head-tuning showed improvement only on Gender but with smaller drop in BLEU. Methods applied during training (Weighting, CoWord Dropout, and Divide and Rule) showed a smaller reduction in BLEU compared to fine-tuning. We attribute this to the smaller discrepancy in the dataset distribution between training and evaluation. Weighting outperformed CoWord Dropout on Gender and Auxiliary. Conversely, CoWord Dropout achieved the highest accuracy on Auxiliary (with Weighting being the second-best) but did not show any improvement for Gender and Formality. Notably, the highest reduction in BLEU was around 1\% compared to the baseline. Lack of improvement exhibited by Adapted Divide and Rule can be attributed to our adaptation implementation, which did not utilize parameter freezing as in the original paper.
Among all methods, metric-based example selection achieved the highest average ctxPro accuracy across phenomena, while token-level loss weighting was the most effective among annotation-based approaches, demonstrating that both proposed techniques can substantially improve context utilization.

\subsection{Multilingual Results}

We trained models based on NLLB-200 600M on all relevant language-directions using annotation-free methods \change{(due to the lack of exhaustive annotations on the OpenSubtitles dataset; see Table~\ref{tab:methods-details})} to assess their performance in the multilingual setting. For CoWord Dropout, we used the same values of p (0.1, 0.2, and 0.3), and for Metric-based example selection, we set k=10,000 per language direction and the number of epochs equal to 1, 2, and 5. The results 
aggregated over language directions can be seen in Table~\ref{tab:nllb-fine-tuning-results} and extended results in Appendix~\ref{sec:extended-fine-tuning-results}.

Fine-tuning on examples selected by MaxPCXMI outperformed all baselines in terms of ctxPro accuracy across phenomena, with the highest improvement of 5.8, 1.9, and 1.6 percentage points (on average) for Auxiliary, Gender, and Animacy, respectively. Contrary to the English-to-German experiments, no improvement (on average) was observed for Formality. This was caused by a drop of up to 1 percentage point in the English-to-French direction, which offsets small gains in other language directions. 
These accuracy improvements came at the cost of a greater reduction in BLEU compared to other methods, and both trends—accuracy gains and BLEU drops—intensified with more fine-tuning epochs, mirroring the patterns seen in the single-language-direction experiments. \change{It should be noted that MaxPCXMI was effectively trained for more updates than other methods in this experiment but additional training did not improve their results (as can be seen in Appendix~\ref{sec:extended-fine-tuning-results}).}

\section{Conclusions}

This work provided a systematic empirical evaluation of the influence of training data composition, in terms of contextually rich examples, on the context utilization capabilities for MT models. By systematically adapting the proportion of contextually rich examples in the training data, we demonstrated that such data sparsity is the key bottleneck in learning to leverage context efficiently. Crucially, we found that (1) models do not generalize well across different contextual phenomena (e.g. gender or formality) and (2) while there is some cross-lingual transfer, it was not significantly higher between languages in the same linguistic sub-family.

Motivated by these findings, we proposed two methods designed to mitigate the effect of data sparsity in context-aware MT: token-level loss weighting (based on token-level annotations of context-dependent words) and metric-based instance selection (fine-tuning on most contextually important examples). Both methods significantly improved context utilization without the need for extensive architectural changes or additional annotated data. Notably, the metric based method showed strong gains across multiple phenomena and language directions. 

In practical terms, data composition and targeted training should be considered as potential solutions to developing strong context-aware MT models.
In future work, combine the strengths of weighting and metric-based example selection.


\section{Limitations}

While we investigate many language directions and three sub-families, all of them come from the Indo-European family. This limitation was imposed by the language directions covered by ctxPro toolset. Additionally, for the single language pair setting, we only tested English-to-German direction. We suspect that the uncovered effects of data composition go beyond the tested language pairs, but this claim has not been tested experimentally.

For encoder-decoder architectures, we only tested the single encoder approach (standard Transformer) and multi-encoder models lay beyond the scope of this study.
For decoder-only (LLM) setting, we based our experiments on a single model (Towerbase 7B). Both different model sizes and families could exhibit different behaviors. Furthermore, we tested the proposed methods for enhancing context utilizaiton only on the encoder-decoder models.  

\section*{Acknowledgments}
The research presented in this paper was conducted as part of VOXReality project\footnote{\url{https://voxreality.eu/}}, which was funded by the European Union Horizon Europe program under grant agreement No 101070521.
This work used the Dutch national e-infrastructure with the support of the SURF Cooperative using grant no. EINF-12385.

\bibliography{custom,anthology-1,anthology-2}

\appendix

\section{Details of ctxPro Dataset}
\label{sec:phenomena}

\change{In this section, we provide a short description of the context-dependent phenomena that can be identified by the ctxPro toolset \citep{wicks-post-2023-identifying}:}
\begin{itemize}[topsep=3pt,itemsep=3pt,partopsep=0pt, parsep=0pt, itemindent=15pt, leftmargin=0pt]
    \item \textbf{Gender} (anaphoric pronouns) - translating a pronoun from a non-gendered language to a language with gendered nouns. Available for English-to-X language directions, where X is \{German, French, Polish, Russian, and Spanish\}.

    \item \textbf{Formality} (anaphoric pronouns) - translating into a language with different second-person pronouns distinguishing intimate from formal relationships between speakers from a language lacking this distinction. Available for English-to-X language directions, where X is \{German, French, Polish, Russian, and Spanish\}.
    
    \item \textbf{Animacy} (anaphoric pronouns) - translating into English, a language that distinguishes between animate (she/he) and inanimate (it) pronouns, from a language that does not exhibit this distinction. Available for X-to-English language directions, where X is \{German, French, Polish, Russian, and Spanish\}.

    \item \textbf{Auxiliary} (verb phrase ellipsis) - translating into a language that require the head of the verb phrase from a language that allows for only the modal or auxiliary to be used. Available for English-to-X language directions, where X is \{German, French, Polish, Russian, and Spanish\}.
    
    \item \textbf{Inflection} (verb phrase ellipsis) - translating into a language with noun morphology dependent on the grammatical role from a language where this is not the case. Available for English-to-Polish and English-to-Russian language directions.
    
\end{itemize}

In Table~\ref{tab:ctxpro-distances} we present the number of examples in the ctxPro dataset with a particular antecedent distance. Additionally, we present the proportion of examples that have the antecedent distance larger than three, which is beyond the context size available to our models. Note that for Formality, the antecedent distances are not specified. We refer the reader to the original paper \citep{wicks-post-2023-identifying} for more details.

\begin{table*}[ht]
    \centering
    \begin{tabular}{llrrrrrr}
    \hline
    \textbf{Direction} & \textbf{Phenomenon} & \multicolumn{6}{c}{\textbf{Antecedent Distance}} \\
     &  & \textbf{0} & \textbf{1} & \textbf{2} & \textbf{3} & \textbf{>3} & \textbf{\% >3} \\
    \hline
    En$\leftrightarrow$De & Auxiliary & 0 & 1754 & 498 & 256 & 672 & 21\% \\
     & Gender & 7307 & 13731 & 4814 & 2308 & 3480 & 11\% \\
     & Animacy & 5309 & 9493 & 3115 & 1362 & 1956 & 9\% \\
    \hline
    En$\leftrightarrow$Es & Auxiliary & 0 & 4922 & 1051 & 323 & 664 & 10\% \\
     & Gender & 4126 & 6979 & 2702 & 1317 & 2392 & 14\% \\
     & Animacy & 2852 & 4102 & 1550 & 750 & 1291 & 12\% \\
    \hline
    En$\leftrightarrow$Fr & Auxiliary & 0 & 5263 & 1327 & 474 & 1258 & 15\% \\
     & Gender & 11236 & 18037 & 6294 & 2921 & 4887 & 11\% \\
     & Animacy & 5468 & 8350 & 2873 & 1317 & 1992 & 10\% \\
    \hline
    En$\leftrightarrow$It & Auxiliary & 0 & 3590 & 1018 & 344 & 972 & 16\% \\
     & Gender & 6117 & 7128 & 2630 & 1365 & 2173 & 11\% \\
     & Animacy & 3277 & 3708 & 1367 & 707 & 1057 & 10\% \\
    \hline
    En$\leftrightarrow$Pl & Auxiliary & 0 & 5437 & 1180 & 391 & 1077 & 13\% \\
     & Gender & 17186 & 25242 & 8201 & 3906 & 5992 & 10\% \\
     & inflection & 0 & 12905 & 5094 & 3235 & 8766 & 29\% \\
     & Animacy & 3455 & 5565 & 1784 & 855 & 1245 & 10\% \\
    \hline
    En$\leftrightarrow$Ru & Auxiliary & 0 & 6056 & 1467 & 402 & 742 & 9\% \\
     & Gender & 8227 & 14283 & 4873 & 2243 & 3322 & 10\% \\
     & inflection & 0 & 15042 & 4659 & 2746 & 7553 & 25\% \\
     & Animacy & 5460 & 9760 & 3422 & 1565 & 2323 & 10\% \\
    \hline
    \end{tabular}
    \caption{Number of examples in ctxPro dataset with certain values for antecedent distance for used language directions and phenomena. Antecedent distances larger than 3 were combined and we also show the proportion of those examples in the dataset. Note that for Formality, the antecedent distance is not specified.}
    \label{tab:ctxpro-distances}
\end{table*}

\section{Composition of the Datasets}
\label{sec:datasets-composition}
In this section, we describe how the constructed datasets were created.
\change{Table~\ref{tab:dense-datasets-composition} shows the sizes of the dense component datasets.}
For the \textit{Pure IWSLT} setting, we start with the IWSLT-sparse (123,000 examples with no annotations) and progressively replace it with the examples sampled from IWSLT-dense. The steps are based on the size of the IWSLT-dense dataset for a particular phenomenon: 3,000 and 6,915 (full size) for Gender, 10,000 and 21,977 (full size) for Formality, and 19 (full size) for Auxiliary.
For the \textit{IWSLT + OS} setting, we start with the datasets formed by combining IWSLT-sparse with examples sampled from OS-rand. To maximize the density of the resulting datasets, we set the number of examples sampled from OS-rand to be dependent on the phenomenon and equal to the (rounded) size of the OS-dense datasets: 12,000 for Gender, 17,000 for Formality, and 1,200 for Auxiliary. We start by replacing examples from IWSLT-sparse (we retain the steps from the Pure IWSLT setting). After reaching the maximum density in the IWSLT portion of the dataset, we start replacing OS-rand with OS-dense in the following steps: 4,000, 8,000, and 12,000 for Gender, 6,000, 12,000, and 17,000 for Formality, and 400, 800, and 1,200 for Auxiliary.

Tables~\ref{tab:gender-dataset-compositions}, \ref{tab:formality-dataset-compositions}, and \ref{tab:auxiliary-dataset-compositions} show the composition of the training datasets we used in the experiments for Gender, Formality, and Auxiliary phenomena, respectively. Each example was encoded with the context size ranging from zero to the maximum context size (three in our experiments), increasing the size of the datasets four times.

In the multilingual experiments, we formed the baseline training dataset by sampling 50,000 examples from OpenSubtitles (OS-rand) for each language direction we considered. For each phenomenon in a language direction, we replaced examples with the rich ones: 6,900 for Gender, 10,000 for Formality, 1,200 for Auxiliary, 10,000 for Inflection, and 4,000 for Animacy.

\begin{table*}[ht]
    \centering
    \begin{tabular}{llrrrrr}
        \hline
        \textbf{Dataset} & \textbf{Language} & \textbf{Gender} & \textbf{Formality} & \textbf{Auxiliary} & \textbf{Inflection} & \textbf{Animacy} \\
        \hline
        IWSLT-dense & En$\rightarrow$De & 6,915 & 21,977 & 19 & - & - \\
        \hline
        OS-dense & En$\leftrightarrow$De & 12,326 & 16,064 & 1,230 & - & 8,334 \\
         & En$\leftrightarrow$Es & 6,936 & 20,374 & 2,768 & - & 4,211 \\
         & En$\leftrightarrow$Fr & 16,804 & 10,858 & 3,314 & - & 7,904 \\
         & En$\leftrightarrow$Pl & 23,683 & 41,806 & 3,184 & 10,897 & 5,112 \\
         & En$\leftrightarrow$Ru & 8,141 & 14,211 & 3,443 & 10,971 & 4,237 \\
        \hline
    \end{tabular}
    \caption{Sizes of the dense component datasets divided into phenomena (columns).}
    \label{tab:dense-datasets-composition}
\end{table*}

\begin{table*}[ht]
    \centering
    \begin{tabular}{lrrrrr}
        \hline
        \textbf{Setting}    & \textbf{IWSLT-sparse} & \textbf{IWSLT-dense} & \textbf{OS-rand} & \textbf{OS-dense} & \textbf{Total}   \\
        \hline
        Pure IWSLT & 123,000      & 0           & 0       & 0        & 123,000 \\
         & 120,000      & 3,000       & 0       & 0        & 123,000 \\
         & 116,085      & 6,915       & 0       & 0        & 123,000 \\
        \hline
        IWSLT+OS   & 123,000      & 0           & 12,000  & 0        & 135,000 \\
           & 120,000      & 3,000       & 12,000  & 0        & 135,000 \\
           & 116,085      & 6,915       & 12,000  & 0        & 135,000 \\
           & 116,085      & 6,915       & 8,000   & 4,000    & 135,000 \\
           & 116,085      & 6,915       & 4,000   & 8,000    & 135,000 \\
           & 116,085      & 6,915       & 0       & 12,000   & 135,000 \\
        \hline
    \end{tabular}
    \caption{Number of examples from datasets that were used to compose training datasets (in rows) for the \textbf{Gender} phenomenon in the single language direction (English-to-German) setting.}
    \label{tab:gender-dataset-compositions}
\end{table*}

\begin{table*}[ht]
    \centering
    \begin{tabular}{lrrrrr}
    \hline
    \textbf{Setting}    & \textbf{IWSLT-sparse} & \textbf{IWSLT-dense} & \textbf{OS-rand} & \textbf{OS-dense}& \textbf{Total}   \\ 
    \hline
    Pure IWSLT & 123,000      & 0           & 0       & 0       & 123,000 \\
     & 113,000      & 10,000      & 0       & 0       & 123,000 \\ 
       & 101,023    & 21,977      & 0       & 0       & 123,000 \\ 
    \hline
    IWSLT+OS   & 123,000      & 0           & 17,000     & 0       & 140,000 \\ 
       & 113,000      & 10,000      & 17,000     & 0       & 140,000 \\ 
       & 101,023      & 21,977      & 17,000     & 0       & 140,000 \\ 
       & 101,023      & 21,977      & 11,000     & 6,000      & 140,000 \\ 
       & 101,023      & 21,977      & 5,000      & 12,000     & 140,000 \\ 
       & 101,023      & 21,977      & 0       & 17,000     & 140,000 \\ 
    \hline
    \end{tabular}
    \caption{Number of examples from datasets that were used to compose training datasets (in rows) for the \textbf{Formality} phenomenon in the single language direction (English-to-German) setting.}
    \label{tab:formality-dataset-compositions}
\end{table*}

\begin{table*}[ht]
    \centering
    \begin{tabular}{lrrrrr}
        \hline
        \textbf{Setting}    & \textbf{IWSLT-sparse }& \textbf{IWSLT-dense} & \textbf{OS-rand} & \textbf{OS-dense} & \textbf{Total}   \\ 
        \hline
        Pure IWSLT & 123,000      & 0           & 0       & 0       & 123,000 \\
         & 122,981      & 19          & 0       & 0       & 123,000 \\ 
        \hline
        IWSLT+OS   & 123,000      & 0           & 1,200   & 0       & 124,200 \\ 
           & 122,981      & 19          & 1,200   & 0       & 124,200 \\ 
           & 122,981      & 19          & 800     & 400     & 124,200 \\ 
           & 122,981      & 19          & 400     & 800     & 124,200 \\ 
           & 122,981      & 19          & 0       & 1,200   & 124,200 \\ 
        \hline
    \end{tabular}
    \caption{Number of examples from datasets that were used to compose training datasets (in rows) for the \textbf{Auxiliary} phenomenon in the single language direction (English-to-German) setting.}
    \label{tab:auxiliary-dataset-compositions}
\end{table*}


\section{Details of Context-aware Training}
\label{sec:training-hyperparameters}
We implemented all experiments in \textit{Huggingface transformers} framework \citep{wolf-etal-2020-transformers}.
\change{We trained the models in the following categories: single language direction (OpusMT en-de\footnote{\url{https://huggingface.co/Helsinki-NLP/opus-mt-en-de}}), multilingual (NLLB-200 600M\footnote{\url{https://huggingface.co/facebook/nllb-200-distilled-600M}}), and LLM-based multilingual (Towerbase 7B\footnote{\url{https://huggingface.co/Unbabel/TowerBase-7B-v0.1}}). Additionally, we repeated the experiments in the single language direction setting using randomly initialized OpusMT model for which we performed sentence-level pre-training on the mixture of IWSLT 2017 en-de training subset and randomly sampled 2.5M sentences from WMT 2019 en-de \cite{barrault-etal-2019-findings} training subset.}
\change{We trained the models with Adafactor optimizer \citep{Shazeer2018AdafactorAL} on a single GPU (NVIDIA GeForce RTX 3090 24GB for Opus MT en-de and NVIDIA H100 80GB for NLLB-200 600M and Towerbase 7B). We used LoRA \citep{hu2022lora} to fine-tune Towerbase models. OpusMT en-de contain 163M parameters, NLLB-200 600M contain 615M parameters, and Towerbase 7B contain 6,770M parameters (32M trainable parameters through LoRA). The inputs during training and prompt used for Towerbase models can be seen in Listings~\ref{lst:llm-input} and \ref{lst:llm-prompt} respectively. We calculated loss during training only based on the target language parts of the examples corresponding to the generations of the model.}

\begin{code}[caption={Input template used for training Towerbase models. The number of sentences in context is the same for source and target sides but can vary from example to example. Sentences are separated by the "<sep>" string.}, label={lst:llm-input}]
[src_lang]: [src_ctx] <sep> [src] \n
[tgt_lang]: [tgt_ctx] <sep> [tgt]
\end{code}

\begin{code}[caption={Prompt template used for generation with Towerbase models. The number of context sentences can vary. Sentences are separated by the "<sep>" string.}, label={lst:llm-prompt}]
[src_lang]: [src_ctx] <sep> [src] \n
[tgt_lang]:
\end{code}

\change{The hyper-parameters are presented in Table~\ref{tab:hyper-parameters}. 
We tuned the hyper-parameters (learning rate, batch size, number of epochs) during the preliminary experiments on OpusMT en-de model with context size of one trained on IWSLT 2017 English-to-German dataset. Hyper-parametes for sentence-level pre-training were tuned on WMT 2019 en-de evaluation subset, and on randomly sampled subset of OpenSubtitles en-de dataset for the fine-tuning of Towerbase 7B model.}

\begin{table*}[h!]
    \centering
    \begin{tabular}{lrrrr}
    \hline
        \textbf{Hyper-parameter} & \textbf{Sentence-level} & \textbf{OpusMT} & \textbf{NLLB-200} & \textbf{Towerbase}    \\
         & \textbf{Pre-training} & \textbf{Fine-tuning} & \textbf{Fine-tuning} & \textbf{Fine-tuning}    \\
        \hline
        Optimizer & Adafactor & Adafactor & Adafactor & Adafactor \\
        Learning Rate & 5e-5 & 1e-5 & 1e-5 & 1e-5 \\
        LR Scheduler & Linear & Inverse Sqrt & Inverse Sqrt & Inverse Sqrt \\
        LR Warmup Ratio & 0.0 & 0.1 & 0.1 & 0.1\\
        Weight Decay & 0.01 & 0.01 & 0.01 & 0.01 \\
        Batch Size & 32 & 32\textsuperscript{a} & 32 & 16 \\
        Gradient Accumulation Steps & 16 & 16\textsuperscript{a} & 16 & 4 \\
        Num Epoch & 30 & 10 & 10 & 3 \\
        Precission & fp16 & fp16 & fp16 & bf16 \\
        Seeds & 1,2,3,4,5 & 1,2,3,4,5 & 1 & 1 \\
        Max Length & 512 & 512 & 1024 &  2048 \\
        Max Context Size & - & 3 & 3 & 3 \\
        Beam size & 5 & 5 & 5 & 1\textsuperscript{b} \\
        Lora alpha & - & - & - & 32 \\
        Lora r & - & - & - & 16 \\
        \hline
    \end{tabular}
    \caption{The hyper-parameters of training and fine-tuning.\\
    \textsuperscript{a} For the cases where the CUDA Out Of Memory error occurred, we reduced the batch size to 16 and increased the Gradient Accumulation Steps to 32, keeping the same effective size of the batch.\\
    \textsuperscript{b} For Towerbase models, we use the greedy decoding strategy.}
    \label{tab:hyper-parameters}
\end{table*}


\section{Extended Data Composition Results}
\label{sec:extended-sparsity-results}

\begin{figure*}
    \centering
    \includegraphics[width=1\linewidth]{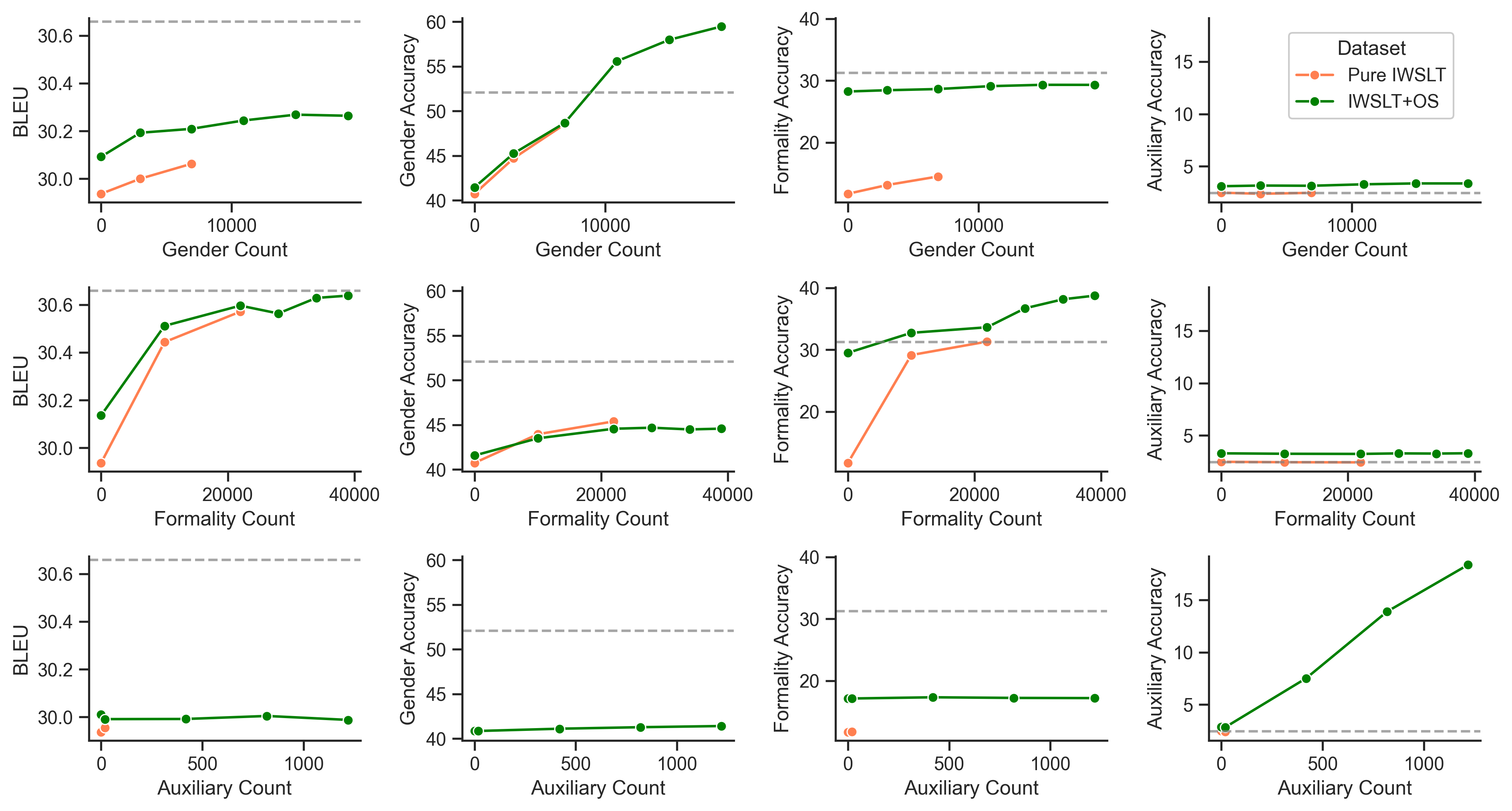}
    \caption{\change{Measured metrics of BLEU on IWSLT 2017 testset, and ctxPro accuracy on Gender, Formality, and Auxiliary phenomena (in columns) of the randomly initialized models trained on the datasets with varying amounts of contextually-rich examples of Gender, Formality, and Auxiliary phenomena (in rows). Shows two experimental settings: Pure IWSLT and combined IWSLT+OS.}}
    \label{fig:sparsity-randinit-results}
\end{figure*}

In this section, we present the extended results of the data composition experiments. For single language pair setting, we measured COMET \citep{rei-etal-2020-comet} (based on \texttt{Unbabel/wmt22-comet-da}) on the IWSLT 2017 en-de testset and evaluated the models on the ContraPro \citep{muller-etal-2018-large} contrastive evaluation. The results for the Pure IWSLT and IWSLT+OS settings can be found in Tables~\ref{tab:pure-iwslt-extended-results} and \ref{tab:iwslt-os-extended-results}, respectively.
The results for English-to-German language direction with models randomly initialized can be seen in Figure~\ref{fig:sparsity-randinit-results}. 

For the multilingual setting, we additionally measured BLEU (we used the sacreBLEU library \citep{post-2018-call} using the default parameters) and COMET on the testsets formed by sampling 20,000 examples from OpenSubtitles 2018 for each language direction. The results for models based on NLLB-200 600M can be seen in Tables~\ref{tab:nllb-bleu-results} and \ref{tab:nllb-comet-results} for BLEU and COMET, respectively. The results for models based on Towerbase 7B can be seen in Tables~\ref{tab:tower-bleu-results} (BLEU) and \ref{tab:towerbase-comet-results} (COMET).

\begin{table}[h]
    \centering
    \begin{tabular}{lrrr}
        \hline
        \textbf{Dataset} & \textbf{Count} & \textbf{COMET} & \textbf{ContraPro} \\
        \hline
        Sparse & 0 & 0.8415 & 69.23 \\
        Gender & 3,000 & 0.8417 & 74.70 \\
         & 6,915 & 0.8417 & 78.45 \\
        Formality & 10,000 & 0.8429 & 69.55 \\
         & 21,977 & 0.8430 & 70.02 \\
        Auxiliary & 19 & 0.8413 & 69.14 \\
        \hline
    \end{tabular}
    \caption{Performance in terms of COMET on IWSLT 2017 en-de testset and ContraPro accuracy for the models \change{based on OpusMT en-de} in the \textbf{Pure IWSLT} setting trained on datasets with different numbers of examples annotated with different phenomena.}
    \label{tab:pure-iwslt-extended-results}
\end{table}

\begin{table}[h]
    \centering
    \begin{tabular}{lrrr}
        \hline
        \textbf{Dataset} & \textbf{Count} & \textbf{COMET} & \textbf{ContraPro} \\
        \hline
        Gender & 0 & 0.8417 & 70.28 \\
         & 3,000 & 0.8417 & 75.03 \\
         & 6,915 & 0.8420 & 78.52 \\
         & 10,915 & 0.8419 & 83.58 \\
         & 14,915 & 0.8418 & 84.77 \\
         & 18,915 & 0.8420 & 85.24 \\
        \hline
        Formality & 0 & 0.8416 & 70.15 \\
         & 10,000 & 0.8426 & 70.59 \\
         & 21,977 & 0.8428 & 71.12 \\
         & 27,977 & 0.8429 & 71.04 \\
         & 33,977 & 0.8429 & 70.85 \\
         & 38,977 & 0.8430 & 71.03 \\
        \hline
        Auxiliary & 0 & 0.8414 & 69.47 \\
         & 19 & 0.8415 & 69.39 \\
         & 419 & 0.8415 & 69.60 \\
         & 819 & 0.8415 & 69.75 \\
         & 1,219 & 0.8416 & 69.79 \\
        \hline
    \end{tabular}
    \caption{Performance in terms of COMET on IWSLT 2017 en-de testset and ContraPro accuracy for the models \change{based on OpusMT en-de} in the \textbf{IWSLT+OS} setting trained on datasets with different numbers of examples annotated with different phenomena.}
    \label{tab:iwslt-os-extended-results}
\end{table}

\begin{table*}[p]
    \centering
    \small
    \begin{tabular}{lrrrrrrrrrr}
    \hline
    \textbf{Model} & \textbf{En-De} & \textbf{En-Es} & \textbf{En-Fr} & \textbf{En-Pl} & \textbf{En-Ru} & \textbf{De-En} & \textbf{Es-En} & \textbf{Fr-En} & \textbf{Pl-En} & \textbf{Ru-En} \\
    \hline
    Baseline & 26.50 & 37.68 & 29.39 & 21.98 & 24.49 & 32.04 & 41.99 & 32.84 & 29.42 & 31.35 \\
    \hline
    \textbf{Gender} &  &  &  &  &  &  &  &  &  &  \\
    En-De & 26.66 & 37.61 & 29.27 & 21.85 & 24.46 & 31.98 & 42.03 & 32.87 & 29.54 & 31.32 \\
    En-Es & 26.88 & 37.60 & 29.33 & 22.12 & 24.52 & 32.03 & 41.96 & 32.86 & 29.46 & 31.36 \\
    En-Fr  & 26.75 & 37.53 & 29.16 & 22.05 & 24.41 & 32.01 & 41.97 & 32.87 & 29.50 & 31.33 \\
    En-Pl  & 26.80 & 37.57 & 29.21 & 21.54 & 24.48 & 32.05 & 42.00 & 32.86 & 29.53 & 31.34 \\
    En-Ru  & 26.78 & 37.60 & 29.56 & 21.91 & 24.45 & 32.01 & 42.04 & 32.81 & 29.52 & 31.41 \\
    \hline
    \textbf{Formality} &  &  &  &  &  &  &  &  &  &  \\
    En-De  & 26.61 & 37.27 & 29.29 & 21.75 & 24.44 & 31.98 & 42.05 & 32.85 & 29.52 & 31.31 \\
    En-Es  & 26.58 & 37.29 & 29.43 & 21.65 & 24.57 & 32.01 & 42.04 & 32.84 & 29.49 & 31.39 \\
    En-Fr  & 26.70 & 37.63 & 29.67 & 21.89 & 24.48 & 32.02 & 41.99 & 32.92 & 29.52 & 31.37 \\
    En-Pl  & 26.62 & 37.38 & 29.44 & 21.83 & 24.35 & 32.03 & 42.00 & 32.88 & 29.44 & 31.23 \\
    En-Ru  & 26.88 & 37.53 & 29.36 & 22.05 & 24.22 & 32.04 & 42.03 & 32.91 & 29.50 & 31.39 \\
    \hline
    \textbf{Auxiliary} &  &  &  &  &  &  &  &  &  &  \\
    En-De  & 26.86 & 37.57 & 29.26 & 21.77 & 24.48 & 32.01 & 42.08 & 32.91 & 29.51 & 31.42 \\
    En-Es  & 26.88 & 37.44 & 29.38 & 22.01 & 24.46 & 32.09 & 41.98 & 32.85 & 29.46 & 31.40 \\
    En-Fr  & 26.94 & 37.53 & 29.56 & 21.97 & 24.42 & 32.01 & 41.99 & 32.83 & 29.51 & 31.28 \\
    En-Pl  & 26.65 & 37.69 & 29.33 & 21.70 & 24.47 & 32.04 & 42.05 & 32.82 & 29.47 & 31.26 \\
    En-Ru  & 26.73 & 37.50 & 29.35 & 22.03 & 24.55 & 32.08 & 41.95 & 32.84 & 29.51 & 31.36 \\
    \hline
    \textbf{Inflection} &  &  &  &  &  &  &  &  &  &  \\
    En-Pl  & 26.95 & 37.58 & 29.41 & 21.68 & 24.59 & 32.07 & 41.98 & 32.87 & 29.49 & 31.40 \\
    En-Ru  & 26.80 & 37.63 & 29.31 & 21.90 & 24.43 & 32.06 & 42.04 & 32.85 & 29.51 & 31.30 \\
    \hline
    \textbf{Animacy} &  &  &  &  &  &  &  &  &  &  \\
    De-En  & 26.80 & 37.43 & 29.32 & 21.84 & 24.65 & 32.05 & 42.05 & 32.84 & 29.48 & 31.41 \\
    Es-En  & 26.83 & 37.59 & 29.39 & 22.20 & 24.50 & 32.02 & 41.97 & 32.81 & 29.51 & 31.27 \\
    Fr-En  & 26.93 & 37.70 & 29.23 & 21.85 & 24.55 & 32.04 & 42.02 & 32.88 & 29.46 & 31.27 \\
    Pl-En  & 26.71 & 37.55 & 29.35 & 21.89 & 24.46 & 32.09 & 42.01 & 32.88 & 29.44 & 31.35 \\
    Ru-En  & 26.83 & 37.51 & 29.35 & 21.73 & 24.48 & 32.00 & 41.95 & 32.86 & 29.48 & 31.35 \\
    \hline
    \end{tabular}
    \caption{BLEU scores for the models \change{based on NLLB-200 600M} trained on datasets with different densities of annotated examples in the multilingual setting on the test subsets of the OpenSubtitles 2018 datasets for all relevant language pairs.}
    \label{tab:nllb-bleu-results}
\end{table*}

\begin{table*}[p]
    \centering
    \small
    \begin{tabular}{lrrrrrrrrrr}
    \hline
    \textbf{Model} & \textbf{En-De} & \textbf{En-Es} & \textbf{En-Fr} & \textbf{En-Pl} & \textbf{En-Ru} & \textbf{De-En} & \textbf{Es-En} & \textbf{Fr-En} & \textbf{Pl-En} & \textbf{Ru-En} \\
    \hline
    Baseline & 0.8023 & 0.8459 & 0.8005 & 0.8171 & 0.8321 & 0.8182 & 0.8522 & 0.8192 & 0.8009 & 0.8086 \\
    \hline
    \textbf{Gender} &  &  &  &  &  &  &  &  &  &  \\
    En-De  & 0.8025 & 0.8456 & 0.8001 & 0.8171 & 0.8325 & 0.8182 & 0.8522 & 0.8189 & 0.8011 & 0.8085 \\
    En-Es  & 0.8025 & 0.8462 & 0.8004 & 0.8172 & 0.8326 & 0.8181 & 0.8521 & 0.8193 & 0.8011 & 0.8086 \\
    En-Fr  & 0.8023 & 0.8456 & 0.8000 & 0.8172 & 0.8322 & 0.8182 & 0.8521 & 0.8192 & 0.8011 & 0.8085 \\
    En-Pl  & 0.8025 & 0.8458 & 0.8004 & 0.8176 & 0.8324 & 0.8182 & 0.8522 & 0.8193 & 0.8011 & 0.8084 \\
    En-Ru  & 0.8021 & 0.8456 & 0.7999 & 0.8168 & 0.8321 & 0.8182 & 0.8523 & 0.8189 & 0.8009 & 0.8086 \\
    \hline
    \textbf{Formality} &  &  &  &  &  &  &  &  &  &  \\
    En-De  & 0.8023 & 0.8456 & 0.8002 & 0.8168 & 0.8324 & 0.8181 & 0.8522 & 0.8190 & 0.8010 & 0.8084 \\
    En-Es  & 0.8026 & 0.8455 & 0.8003 & 0.8171 & 0.8325 & 0.8182 & 0.8523 & 0.8191 & 0.8011 & 0.8087 \\
    En-Fr  & 0.8024 & 0.8458 & 0.8008 & 0.8173 & 0.8321 & 0.8183 & 0.8522 & 0.8192 & 0.8011 & 0.8087 \\
    En-Pl  & 0.8024 & 0.8456 & 0.8005 & 0.8176 & 0.8325 & 0.8185 & 0.8523 & 0.8192 & 0.8009 & 0.8085 \\
    En-Ru  & 0.8022 & 0.8456 & 0.8001 & 0.8171 & 0.8318 & 0.8183 & 0.8524 & 0.8190 & 0.8009 & 0.8085 \\
    \hline
    \textbf{Auxiliary} &  &  &  &  &  &  &  &  &  &  \\
    En-De  & 0.8023 & 0.8458 & 0.8001 & 0.8171 & 0.8321 & 0.8185 & 0.8524 & 0.8190 & 0.8011 & 0.8085 \\
    En-Es  & 0.8024 & 0.8458 & 0.8006 & 0.8174 & 0.8327 & 0.8185 & 0.8522 & 0.8191 & 0.8011 & 0.8085 \\
    En-Fr  & 0.8025 & 0.8455 & 0.7999 & 0.8165 & 0.8322 & 0.8181 & 0.8521 & 0.8189 & 0.8010 & 0.8085 \\
    En-Pl  & 0.8026 & 0.8458 & 0.8001 & 0.8170 & 0.8321 & 0.8183 & 0.8522 & 0.8191 & 0.8009 & 0.8083 \\
    En-Ru  & 0.8024 & 0.8457 & 0.8001 & 0.8169 & 0.8326 & 0.8183 & 0.8520 & 0.8190 & 0.8009 & 0.8085 \\
    \hline
    \textbf{Inflection} &  &  &  &  &  &  &  &  &  &  \\
    En-Pl  & 0.8025 & 0.8458 & 0.8004 & 0.8162 & 0.8323 & 0.8184 & 0.8522 & 0.8191 & 0.8010 & 0.8087 \\
    En-Ru  & 0.8021 & 0.8457 & 0.7999 & 0.8168 & 0.8309 & 0.8184 & 0.8523 & 0.8190 & 0.8010 & 0.8084 \\
    \hline
    \textbf{Animacy} &  &  &  &  &  &  &  &  &  &  \\
    De-En  & 0.8026 & 0.8458 & 0.8003 & 0.8174 & 0.8324 & 0.8184 & 0.8524 & 0.8188 & 0.8010 & 0.8085 \\
    Es-En  & 0.8025 & 0.8459 & 0.8005 & 0.8171 & 0.8328 & 0.8184 & 0.8522 & 0.8191 & 0.8009 & 0.8086 \\
    Fr-En  & 0.8021 & 0.8458 & 0.8000 & 0.8168 & 0.8325 & 0.8181 & 0.8523 & 0.8191 & 0.8008 & 0.8083 \\
    Pl-En  & 0.8021 & 0.8456 & 0.8004 & 0.8171 & 0.8322 & 0.8183 & 0.8522 & 0.8192 & 0.8008 & 0.8085 \\
    Ru-En  & 0.8022 & 0.8455 & 0.8003 & 0.8172 & 0.8321 & 0.8182 & 0.8521 & 0.8189 & 0.8008 & 0.8083 \\
    \hline
    \end{tabular}
    \caption{COMET scores for the models \change{based on NLLB-200 600M} trained on datasets with different densities of annotated examples in the multilingual setting on the test subsets of the OpenSubtitles 2018 datasets for all relevant language pairs.}
    \label{tab:nllb-comet-results}
\end{table*}

\begin{table*}[p]
    \centering
    \small
    \begin{tabular}{lrrrrrrrr}
    \hline
    \textbf{Model} & \textbf{En-De} & \textbf{En-Es} & \textbf{En-Fr} & \textbf{En-Ru} & \textbf{De-En} & \textbf{Es-En} & \textbf{Fr-En} & \textbf{Ru-En} \\
    \hline
    Baseline & 25.93 & 32.78 & 29.45 & 21.56 & 31.06 & 42.23 & 33.84 & 28.40 \\
    \hline
    \textbf{Gender} &  &  &  &  &  &  &  &  \\
    En-De  & 25.81 & 32.83 & 29.09 & 20.81 & 31.72 & 41.94 & 32.74 & 28.13 \\
    En-Es  & 25.37 & 34.02 & 28.95 & 21.22 & 31.12 & 42.61 & 33.53 & 28.18 \\
    En-Fr  & 25.60 & 34.00 & 28.64 & 21.86 & 31.37 & 42.22 & 33.77 & 28.02 \\
    En-Ru  & 24.74 & 32.82 & 28.92 & 22.02 & 30.94 & 42.55 & 33.86 & 27.28 \\
    \hline
    \textbf{Formality} &  &  &  &  &  &  &  &  \\
    En-De  & 25.61 & 33.84 & 29.00 & 20.45 & 31.50 & 42.71 & 33.47 & 28.35 \\
    En-Es  & 25.87 & 33.61 & 28.87 & 22.05 & 31.40 & 41.37 & 33.96 & 29.01 \\
    En-Fr  & 25.46 & 33.63 & 29.35 & 22.10 & 30.86 & 41.40 & 33.78 & 27.55 \\
    En-Ru  & 25.43 & 32.55 & 29.45 & 22.65 & 30.84 & 42.76 & 33.81 & 27.80 \\
    \hline
    \textbf{Auxiliary} &  &  &  &  &  &  &  &  \\
    En-De  & 26.10 & 31.95 & 28.85 & 21.11 & 31.48 & 41.89 & 33.29 & 28.93 \\
    En-Es  & 25.66 & 32.33 & 29.03 & 21.44 & 31.50 & 41.95 & 33.64 & 27.94 \\
    En-Fr  & 25.75 & 33.30 & 29.19 & 21.91 & 31.12 & 42.26 & 33.83 & 28.76 \\
    En-Ru  & 25.60 & 33.71 & 28.96 & 22.24 & 31.77 & 41.57 & 33.89 & 28.30 \\
    \hline
    \textbf{Inflection} &  &  &  &  &  &  &  &  \\
    En-Ru  & 25.52 & 32.71 & 28.72 & 21.56 & 30.73 & 42.55 & 33.80 & 28.11 \\
    \hline
    \textbf{Animacy} &  &  &  &  &  &  &  &  \\
    De-En  & 25.07 & 34.34 & 28.87 & 21.31 & 30.88 & 41.92 & 33.42 & 28.66 \\
    Es-En  & 25.94 & 32.01 & 29.03 & 21.58 & 30.78 & 43.05 & 33.88 & 27.84 \\
    Fr-En  & 25.32 & 32.97 & 29.22 & 22.39 & 31.40 & 41.72 & 33.15 & 28.60 \\
    Ru-En  & 25.48 & 33.04 & 29.15 & 22.23 & 30.26 & 41.85 & 33.48 & 29.34 \\
    \hline
    \end{tabular}
    \caption{BLEU scores for the models \change{based on Towerbase 7B} trained on datasets with different densities of annotated examples in the multilingual setting on the test subsets of the OpenSubtitles 2018 datasets for all relevant language pairs.}
    \label{tab:tower-bleu-results}
\end{table*}

\begin{table*}[p]
    \centering
    \small
    \begin{tabular}{lrrrrrrrr}
    \hline
    \textbf{Model} & \textbf{En-De} & \textbf{En-Es} & \textbf{En-Fr} & \textbf{En-Ru} & \textbf{De-En} & \textbf{Es-En} & \textbf{Fr-En} & \textbf{Ru-En} \\
    \hline
    Baseline & 0.8003 & 0.8407 & 0.7979 & 0.8336 & 0.8186 & 0.8547 & 0.8236 & 0.8106 \\
    \hline
    \textbf{Gender} &  &  &  &  &  &  &  &  \\
    En-De  & 0.8013 & 0.8410 & 0.7981 & 0.8342 & 0.8193 & 0.8548 & 0.8239 & 0.8114 \\
    En-Es  & 0.8000 & 0.8412 & 0.7979 & 0.8340 & 0.8193 & 0.8546 & 0.8235 & 0.8113 \\
    En-Fr  & 0.8003 & 0.8412 & 0.7983 & 0.8340 & 0.8193 & 0.8547 & 0.8237 & 0.8110 \\
    En-Ru  & 0.8005 & 0.8409 & 0.7982 & 0.8348 & 0.8187 & 0.8540 & 0.8236 & 0.8108 \\
    \hline
    \textbf{Formality} &  &  &  &  &  &  &  &  \\
    En-De  & 0.8013 & 0.8409 & 0.7979 & 0.8344 & 0.8191 & 0.8547 & 0.8241 & 0.8114 \\
    En-Es  & 0.8004 & 0.8404 & 0.7978 & 0.8341 & 0.8192 & 0.8544 & 0.8235 & 0.8116 \\
    En-Fr  & 0.8005 & 0.8412 & 0.7988 & 0.8340 & 0.8188 & 0.8546 & 0.8238 & 0.8109 \\
    En-Ru  & 0.8007 & 0.8414 & 0.7979 & 0.8345 & 0.8185 & 0.8542 & 0.8236 & 0.8107 \\
    \hline
    \textbf{Auxiliary} &  &  &  &  &  &  &  &  \\
    En-De  & 0.8005 & 0.8405 & 0.7974 & 0.8339 & 0.8189 & 0.8545 & 0.8239 & 0.8112 \\
    En-Es  & 0.8005 & 0.8405 & 0.7978 & 0.8338 & 0.8188 & 0.8547 & 0.8236 & 0.8114 \\
    En-Fr  & 0.8006 & 0.8406 & 0.7979 & 0.8338 & 0.8189 & 0.8546 & 0.8236 & 0.8109 \\
    En-Ru  & 0.8004 & 0.8408 & 0.7982 & 0.8339 & 0.8190 & 0.8541 & 0.8236 & 0.8107 \\
    \hline
    \textbf{Inflection} &  &  &  &  &  &  &  &  \\
    En-Ru  & 0.8007 & 0.8409 & 0.7982 & 0.8335 & 0.8189 & 0.8546 & 0.8236 & 0.8108 \\
    \hline
    \textbf{Animacy} &  &  &  &  &  &  &  &  \\
    De-En  & 0.8002 & 0.8405 & 0.7978 & 0.8342 & 0.8190 & 0.8548 & 0.8231 & 0.8112 \\
    Es-En  & 0.8007 & 0.8405 & 0.7977 & 0.8340 & 0.8192 & 0.8550 & 0.8237 & 0.8110 \\
    Fr-En  & 0.8003 & 0.8407 & 0.7981 & 0.8337 & 0.8189 & 0.8544 & 0.8236 & 0.8109 \\
    Ru-En  & 0.8004 & 0.8409 & 0.7978 & 0.8337 & 0.8191 & 0.8545 & 0.8237 & 0.8111 \\
    \hline
    \end{tabular}
    \caption{COMET scores for the models \change{based on Towerbase 7B} trained on datasets with different densities of annotated examples in the multilingual setting on the test subsets of the OpenSubtitles 2018 datasets for all relevant language pairs.}
    \label{tab:towerbase-comet-results}
\end{table*}

\section{Extended Fine-tuning Results}
\label{sec:extended-fine-tuning-results}

For the English-to-German experiment, apart from BLEU and ctxPro accuracy, we also measured COMET \citep{rei-etal-2020-comet} (based on \texttt{Unbabel/wmt22-comet-da}) on the IWSLT 2017 en-de testset and the accuracy on the ContraPro contrastive evaluation. The results (including BLEU and ctxPro accuracies) can be seen in Table~\ref{tab:fine-tuning-extended-results}.

\begin{table*}[h!]
    \centering
    \begin{tabular}{lrrrrrr}
        \hline
        \textbf{Model} & \textbf{BLEU} & \textbf{COMET} & \textbf{Gender} & \textbf{Formality} & \textbf{Auxiliary} & \textbf{ContraPro} \\
        \hline
        Baseline & 33.93 & 0.8431 & 60.52\% & 38.63\% & 6.81\% & 78.88\% \\
        \hline
        Fine-tuning e=1 & 33.60 & 0.8416 & 66.79\% & 39.30\% & 6.30\% & 83.02\% \\
        Fine-tuning e=2 & 33.59 & 0.8416 & 67.49\% & 39.34\% & 6.37\% & 83.78\% \\
        Fine-tuning e=5 & 33.60 & 0.8415 & 68.20\% & 39.49\% & 6.48\% & 84.50\% \\
        Head-tuning h=1 & 33.89 & 0.8428 & 63.28\% & 38.64\% & 6.43\% & 82.61\% \\
        Head-tuning h=2 & 33.85 & 0.8427 & 64.04\% & 38.58\% & 6.44\% & 83.40\% \\
        Head-tuning h=3 & 33.80 & 0.8425 & 64.75\% & 38.27\% & 6.45\% & 84.36\% \\
        Weighting $\lambda$=2 & 33.94 & 0.8430 & 64.35\% & 39.14\% & 7.18\% & 83.10\% \\
        Weighting $\lambda$=5 & 33.83 & 0.8430 & 65.72\% & 39.48\% & 7.67\% & 84.63\% \\
        Weighting $\lambda$=10 & 33.74 & 0.8426 & 66.24\% & 39.81\% & 8.10\% & 85.11\% \\
        \hline
        Adapted D\&R None & 33.95 & 0.8429 & 60.77\% & 38.17\% & 7.01\% & 78.66\% \\
        CoWord p=0.1 & 33.98 & 0.8435 & 60.54\% & 38.72\% & 7.79\% & 78.65\% \\
        CoWord p=0.2 & 33.95 & 0.8436 & 60.47\% & 38.72\% & 8.22\% & 78.52\% \\
        CoWord p=0.3 & 33.88 & 0.8433 & 60.29\% & 38.68\% & 8.59\% & 78.39\% \\
        MaxPCXMI e=1 & 33.71 & 0.8420 & 66.16\% & 41.11\% & 6.84\% & 82.95\% \\
        MaxPCXMI e=2 & 33.70 & 0.8418 & 66.86\% & 41.44\% & 6.99\% & 83.79\% \\
        MaxPCXMI e=5 & 33.62 & 0.8414 & 67.31\% & 41.82\% & 7.18\% & 84.39\% \\
        \hline
    \end{tabular}
    \caption{Performance in terms of BLEU and COMET on IWSLT 2017 en-de testset and ctxPro and ContraPro accuracy for the different methods \change{applied to OpusMT en-de model}. Number of epochs is noted as "e", and CoWord Dropout probability as "p", number of tuned heads as "h", and weighting strength as "$\lambda$".}
    \label{tab:fine-tuning-extended-results}
\end{table*}

\begin{figure*}[h]
    \centering
    \includegraphics[width=1\linewidth]{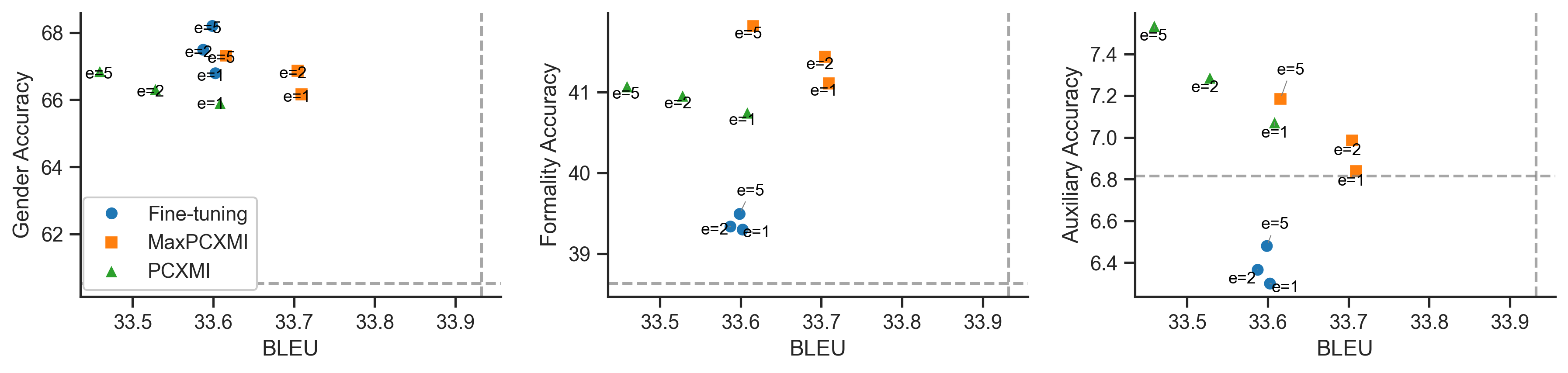}
    \caption{Accuracy of ctxPro English-to-German phenomena (Gender, Formality, and Auxiliary) against BLEU on the IWSLT 2017 en-de testset of the fine-tuned models with Metric-based (PCXMI and MaxPCXMI) and annotation-based (for comparison) selection of examples. \change{Models are based on OpusMT en-de}. Labels show the number of epochs ("e").}
    \label{fig:finetuning-pcxmi-opusmt}
\end{figure*}

Next, we present the results of Metric-based selection of examples for fine-tuning for two metrics: PCXMI \citep{fernandes-etal-2023-translation} and MaxPCXMI (ours). We fine-tuned the models for 1, 2, and 5 epochs and repeated the experiment 5 times with different seeds (using the base context-aware model trained with the corresponding seed). The averaged results can be seen in Figure~\ref{fig:finetuning-pcxmi-opusmt}. Selecting examples based on MaxPCXMI outperforms PCXMI in Gender and Formality at a lower reduction in BLEU. PCXMI achieves a better increase in Auxiliary but reduces BLEU even below the level of the annotation-based method.

\begin{figure*}[h]
    \centering
    \includegraphics[width=1\linewidth]{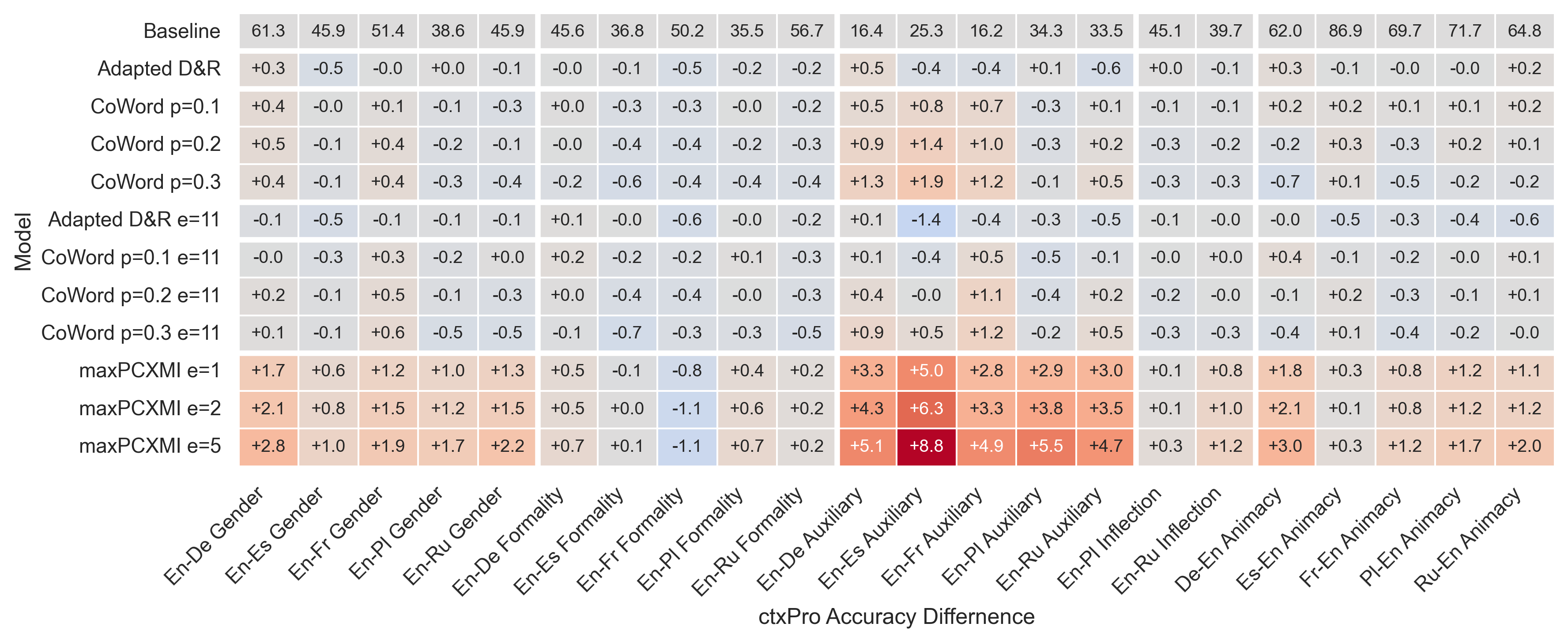}
    \caption{Measured ctxPro accuracy on all phenomena for each of the relevant language directions (in columns) of tested methods (in rows) \change{applied to NLLB-200 600M model}.}
    \label{fig:fine-tuning-nllb-results}
\end{figure*}

The un-aggregated results of the trained models for each language direction in the multilingual experiment can be seen in Figure~\ref{fig:fine-tuning-nllb-results} (including models trained for one more epoch) and Tables~\ref{tab:nllb-fine-tuning-bleu-results} and \ref{tab:nllb-fine-tuning-comet-results} for ctxPro accuracies, BLEU and COMET, respectively.

\begin{table*}[h!]
    \centering
    \small
    \begin{tabular}{lrrrrrrrrrr}
    \hline
    \textbf{Model} & \textbf{En-De} & \textbf{En-Es} & \textbf{En-Fr} & \textbf{En-Pl} & \textbf{En-Ru} & \textbf{De-En} & \textbf{Es-En} & \textbf{Fr-En} & \textbf{Pl-En} & \textbf{Ru-En} \\
    \hline
    Baseline & 26.50 & 37.68 & 29.39 & 21.98 & 24.49 & 32.04 & 41.99 & 32.84 & 29.42 & 31.35 \\
    \hline
    Adapted D\&R & 26.50 & 37.00 & 29.48 & 22.00 & 24.44 & 32.05 & 42.01 & 32.88 & 29.50 & 31.30 \\
    CoWord p=0.1 & 26.72 & 37.48 & 28.86 & 21.89 & 24.27 & 32.10 & 41.97 & 32.77 & 29.41 & 31.31 \\
    CoWord p=0.2 & 26.45 & 37.31 & 29.27 & 22.01 & 24.25 & 32.05 & 41.88 & 32.75 & 29.35 & 31.30 \\
    CoWord p=0.3 & 26.58 & 37.61 & 29.48 & 21.95 & 24.15 & 32.11 & 41.82 & 32.68 & 29.28 & 31.22 \\
    \hline
    MaxPCXMI e=1 & 26.00 & 37.04 & 28.71 & 21.23 & 24.02 & 31.89 & 41.78 & 32.73 & 29.35 & 30.76 \\
    MaxPCXMI e=2 & 26.04 & 37.02 & 28.59 & 21.34 & 23.90 & 31.81 & 41.81 & 32.71 & 29.31 & 30.68 \\
    MaxPCXMI e=5 & 26.09 & 36.93 & 28.74 & 21.29 & 23.85 & 31.78 & 41.65 & 32.62 & 29.22 & 30.46 \\
    \hline
    \end{tabular}
    \caption{BLEU scores for the methods \change{applied to NLLB-200 600M model} in the multilingual setting on the test subsets of the OpenSubtitles 2018 datasets for all relevant language pairs.}
    \label{tab:nllb-fine-tuning-bleu-results}
\end{table*}

\begin{table*}[h!]
    \centering
    \small
    \begin{tabular}{lrrrrrrrrrr}
    \hline
    \textbf{Model} & \textbf{En-De} & \textbf{En-Es} & \textbf{En-Fr} & \textbf{En-Pl} & \textbf{En-Ru} & \textbf{De-En} & \textbf{Es-En} & \textbf{Fr-En} & \textbf{Pl-En} & \textbf{Ru-En} \\
    \hline
    Baseline & 0.8023 & 0.8459 & 0.8005 & 0.8171 & 0.8321 & 0.8182 & 0.8522 & 0.8192 & 0.8009 & 0.8086 \\
    \hline
    Adapted D\&R & 0.8026 & 0.8456 & 0.8000 & 0.8175 & 0.8322 & 0.8183 & 0.8522 & 0.8191 & 0.8011 & 0.8085 \\
    CoWord p=0.1 & 0.8023 & 0.8454 & 0.7994 & 0.8167 & 0.8317 & 0.8182 & 0.8521 & 0.8188 & 0.8006 & 0.8086 \\
    CoWord p=0.2 & 0.8015 & 0.8453 & 0.7994 & 0.8166 & 0.8316 & 0.8178 & 0.8518 & 0.8187 & 0.8002 & 0.8083 \\
    CoWord p=0.3 & 0.8014 & 0.8453 & 0.7990 & 0.8164 & 0.8313 & 0.8176 & 0.8516 & 0.8183 & 0.7996 & 0.8083 \\
    \hline
    MaxPCXMI e=1 & 0.7990 & 0.8433 & 0.7963 & 0.8125 & 0.8296 & 0.8155 & 0.8501 & 0.8170 & 0.7988 & 0.8057 \\
    MaxPCXMI e=2 & 0.7987 & 0.8431 & 0.7958 & 0.8123 & 0.8296 & 0.8150 & 0.8499 & 0.8167 & 0.7982 & 0.8053 \\
    MaxPCXMI e=5 & 0.7974 & 0.8427 & 0.7947 & 0.8109 & 0.8285 & 0.8137 & 0.8490 & 0.8158 & 0.7970 & 0.8043 \\
    \hline
    \end{tabular}
    \caption{COMET scores for the methods \change{applied to NLLB-200 600M model} in the multilingual setting on the test subsets of the OpenSubtitles 2018 datasets for all relevant language pairs.}
    \label{tab:nllb-fine-tuning-comet-results}
\end{table*}

\section{Statistical Significance}
\label{sec:signifficance}

\change{In this section, we calculate the statistical significance of the fine-tuning results on the single language pair setting. In particular, we employ the paired bootstrap resampling method \citep{koehn-2004-statistical} to calculate whether the differences in obtained results between tested methods are statistically significant. We use sacreBLEU \citep{post-2018-call} implementation extended to other metrics. To include the runs with all seeds, we concatenate the predictions (as well as references) for all runs of a particular model. The results in terms of p-values of the paired bootstrapping of the results are presented in Tables~\ref{tab:fine-tuning-signifficance-bleu} to \ref{tab:fine-tuning-signifficance-auxiliary} for: BLEU on the IWSLT 2017 en-de testset, COMET on the IWSLT 2017 en-de testset, ctxPro Gender accuracy, ctxPro Formality accuracy, and ctxPro Auxiliary accuracy, respectively.}

\begin{table*}[h]
    \centering
    \scriptsize
    \caption{The p-values of the paired bootstrapping of the results in terms of BLEU on the IWSLT2017 English-to-German testset for each pair of the models based on OpusMT en-de. Values <0.05 are in bold.}
    \begin{tabular}{l|cccccccccccccc}
        \hline
        \textbf{Model} & \rotatebox[origin=c]{90}{Baseline} & \rotatebox[origin=c]{90}{CoWord p=0.1} & \rotatebox[origin=c]{90}{CoWord p=0.2} & \rotatebox[origin=c]{90}{CoWord p=0.3} & \rotatebox[origin=c]{90}{Adapted D\&R} & \rotatebox[origin=c]{90}{Fine-tuning e=1} & \rotatebox[origin=c]{90}{Fine-tuning e=2} & \rotatebox[origin=c]{90}{Fine-tuning e=5} & \rotatebox[origin=c]{90}{MaxPCXMI e=1} & \rotatebox[origin=c]{90}{MaxPCXMI e=2} & \rotatebox[origin=c]{90}{ MaxPCXMI e=5 } & \rotatebox[origin=c]{90}{Weighting $\lambda$=2} & \rotatebox[origin=c]{90}{Weighting $\lambda$=5} & \rotatebox[origin=c]{90}{Weighting $\lambda$=10} \\
        \hline
        Baseline  & - & \textbf{\textbf{0.045}} & 0.243 & 0.103 & 0.192 & \textbf{0.001} & \textbf{0.001} & \textbf{0.001} & \textbf{0.001} & \textbf{0.001} & \textbf{0.001} & 0.319 & \textbf{0.001} & \textbf{0.001} \\
        CoWord p=0.1 & \textbf{0.045} & -        & 0.085 & \textbf{0.007} & 0.124 & \textbf{0.001} & \textbf{0.001} & \textbf{0.001} & \textbf{0.001} & \textbf{0.001} & \textbf{0.001} & 0.069 & \textbf{0.001} & \textbf{0.001} \\
        CoWord p=0.2 & 0.243 & 0.085 & - & \textbf{0.018} & 0.372 & \textbf{0.001} & \textbf{0.001} & \textbf{0.001} & \textbf{0.001} & \textbf{0.001} & \textbf{0.001} & 0.281 & \textbf{0.001} & \textbf{0.001} \\
        CoWord p=0.3 & 0.103 & \textbf{0.007} & \textbf{0.018} & - & \textbf{0.049} & \textbf{0.001} & \textbf{0.001} & \textbf{0.001} & \textbf{0.001} & \textbf{0.001} & \textbf{0.001} & 0.080 & 0.107 & \textbf{0.002} \\
        Adapted D\&R & 0.192 & 0.124 & 0.372 & \textbf{0.049} & - & \textbf{0.001} & \textbf{0.001} & \textbf{0.001} & \textbf{0.001} & \textbf{0.001} & \textbf{0.001} & 0.243 & \textbf{0.002} & \textbf{0.001} \\
        Fine-tuning e=1 & \textbf{0.001} & \textbf{0.001} & \textbf{0.001} & \textbf{0.001} & \textbf{0.001} & - & 0.164 & 0.340 & \textbf{0.001} & \textbf{0.001} & 0.244 & \textbf{0.001} & \textbf{0.001} & \textbf{0.001} \\
        Fine-tuning e=2 & \textbf{0.001} & \textbf{0.001} & \textbf{0.001} & \textbf{0.001} & \textbf{0.001} & 0.164 & - & 0.217 & \textbf{0.001} & \textbf{0.002} & 0.138 & \textbf{0.001} & \textbf{0.001} & \textbf{0.001} \\
        Fine-tuning e=5 & \textbf{0.001} & \textbf{0.001} & \textbf{0.001} & \textbf{0.001} & \textbf{0.001} & 0.340 & 0.217 & - & \textbf{0.002} & \textbf{0.002} & 0.224 & \textbf{0.001} & \textbf{0.001} & \textbf{0.001} \\
        MaxPCXMI e=1 & \textbf{0.001} & \textbf{0.001} & \textbf{0.001} & \textbf{0.001} & \textbf{0.001} & \textbf{0.001} & \textbf{0.001} & \textbf{0.002} & - & 0.321 & \textbf{0.001} & \textbf{0.001} & \textbf{0.005} & 0.161 \\
        MaxPCXMI e=2 & \textbf{0.001} & \textbf{0.001} & \textbf{0.001} & \textbf{0.001} & \textbf{0.001} & \textbf{0.001} & \textbf{0.002} & \textbf{0.002} & 0.321 & - & \textbf{0.001} & \textbf{0.001} & \textbf{0.002} & 0.143 \\
        MaxPCXMI e=5 & \textbf{0.001} & \textbf{0.001} & \textbf{0.001} & \textbf{0.001} & \textbf{0.001} & 0.243 & 0.138 & 0.224 & \textbf{0.001} & \textbf{0.001} & - & \textbf{0.001} & \textbf{0.001} & \textbf{0.003} \\
        Weighting $\lambda$=2 & 0.319 & 0.069 & 0.281 & 0.080 & 0.243 & \textbf{0.001} & \textbf{0.001} & \textbf{0.001} & \textbf{0.001} & \textbf{0.001} & \textbf{0.001} & - & \textbf{0.001} & \textbf{0.001} \\
        Weighting $\lambda$=5 & \textbf{0.001} & \textbf{0.001} & \textbf{0.001} & 0.107 & \textbf{0.002} & \textbf{0.001} & \textbf{0.001} & \textbf{0.001} & \textbf{0.005} & \textbf{0.002} & \textbf{0.001} & \textbf{0.001} & - & \textbf{0.001} \\
        Weighting $\lambda$=10 & \textbf{0.001} & \textbf{0.001} & \textbf{0.001} & \textbf{0.002} & \textbf{0.001} & \textbf{0.001} & \textbf{0.001} & \textbf{0.001} & 0.161 & 0.143 & \textbf{0.003} & \textbf{0.001} & \textbf{0.001} & - \\
        \hline
    \end{tabular}
    \label{tab:fine-tuning-signifficance-bleu}
\end{table*}

\begin{table*}[h]
    \centering
    \scriptsize
    \caption{The p-values of the paired bootstrapping of the results in terms of COMET on the IWSLT2017 English-to-German testset for each pair of the models based on OpusMT en-de. Values <0.05 are in bold.}
    \begin{tabular}{l|cccccccccccccc}
        \hline
        \textbf{Model} & \rotatebox[origin=c]{90}{Baseline} & \rotatebox[origin=c]{90}{CoWord p=0.1} & \rotatebox[origin=c]{90}{CoWord p=0.2} & \rotatebox[origin=c]{90}{CoWord p=0.3} & \rotatebox[origin=c]{90}{Adapted D\&R} & \rotatebox[origin=c]{90}{Fine-tuning e=1} & \rotatebox[origin=c]{90}{Fine-tuning e=2} & \rotatebox[origin=c]{90}{Fine-tuning e=5} & \rotatebox[origin=c]{90}{MaxPCXMI e=1} & \rotatebox[origin=c]{90}{MaxPCXMI e=2} & \rotatebox[origin=c]{90}{ MaxPCXMI e=5 } & \rotatebox[origin=c]{90}{Weighting $\lambda$=2} & \rotatebox[origin=c]{90}{Weighting $\lambda$=5} & \rotatebox[origin=c]{90}{Weighting $\lambda$=10} \\
        \hline
        Baseline & - & \textbf{0.003} & \textbf{0.006} & 0.122 & \textbf{0.050} & \textbf{0.001} & \textbf{0.001} & \textbf{0.001} & \textbf{0.001} & \textbf{0.001} & \textbf{0.001} & 0.138 & 0.089 & \textbf{0.002} \\
        CoWord p=0.1 & \textbf{0.003} & - & 0.291 & 0.065 & \textbf{0.001} & \textbf{0.001} & \textbf{0.001} & \textbf{0.001} & \textbf{0.001} & \textbf{0.001} & \textbf{0.001} & \textbf{0.001} & \textbf{0.001} & \textbf{0.001} \\
        CoWord p=0.2 & \textbf{0.006} & 0.291 & - & \textbf{0.009} & \textbf{0.001} & \textbf{0.001} & \textbf{0.001} & \textbf{0.001} & \textbf{0.001} & \textbf{0.001} & \textbf{0.001} & \textbf{0.002} & \textbf{0.001} & \textbf{0.001} \\
        CoWord p=0.3 & 0.122 & 0.065 & \textbf{0.009} & - & \textbf{0.015} & \textbf{0.001} & \textbf{0.001} & \textbf{0.001} & \textbf{0.001} & \textbf{0.001} & \textbf{0.001} & 0.066 & \textbf{0.043} & \textbf{0.001} \\
        Adapted D\&R & \textbf{0.050} & \textbf{0.001} & \textbf{0.001} & \textbf{0.015} & - & \textbf{0.001} & \textbf{0.001} & \textbf{0.001} & \textbf{0.001} & \textbf{0.001} & \textbf{0.001} & 0.127 & 0.191 & 0.068 \\
        Fine-tuning e=1 & \textbf{0.001} & \textbf{0.001} & \textbf{0.001} & \textbf{0.001} & \textbf{0.001} & - & 0.317 & 0.141 & \textbf{0.002} & \textbf{0.043} & 0.087 & \textbf{0.001} & \textbf{0.001} & \textbf{0.001} \\
        Fine-tuning e=2 & \textbf{0.001} & \textbf{0.001} & \textbf{0.001} & \textbf{0.001} & \textbf{0.001} & 0.317 & - & 0.103 & \textbf{0.007} & 0.064 & 0.072 & \textbf{0.001} & \textbf{0.001} & \textbf{0.001} \\
        Fine-tuning e=5 & \textbf{0.001} & \textbf{0.001} & \textbf{0.001} & \textbf{0.001} & \textbf{0.001} & 0.141 & 0.103 & - & \textbf{0.002} & \textbf{0.011} & 0.238 & \textbf{0.001} & \textbf{0.001} & \textbf{0.001} \\
        MaxPCXMI e=1 & \textbf{0.001} & \textbf{0.001} & \textbf{0.001} & \textbf{0.001} & \textbf{0.001} & \textbf{0.002} & \textbf{0.007} & \textbf{0.002} & - & \textbf{0.038} & \textbf{0.001} & \textbf{0.001} & \textbf{0.001} & \textbf{0.003} \\
        MaxPCXMI e=2 & \textbf{0.001} & \textbf{0.001} & \textbf{0.001} & \textbf{0.001} & \textbf{0.001} & \textbf{0.043} & 0.064 & \textbf{0.011} & \textbf{0.038} & - & \textbf{0.001} & \textbf{0.001} & \textbf{0.001} & \textbf{0.001} \\
        MaxPCXMI e=5 & \textbf{0.001} & \textbf{0.001} & \textbf{0.001} & \textbf{0.001} & \textbf{0.001} & 0.087 & 0.072 & 0.238 & \textbf{0.001} & \textbf{0.001} & - & \textbf{0.001} & \textbf{0.001} & \textbf{0.001} \\
        Weighting $\lambda$=2 & 0.138 & \textbf{0.001} & \textbf{0.002} & 0.066 & 0.127 & \textbf{0.001} & \textbf{0.001} & \textbf{0.001} & \textbf{0.001} & \textbf{0.001} & \textbf{0.001} & - & 0.170 & \textbf{0.001} \\
        Weighting $\lambda$=5 & 0.089 & \textbf{0.001} & \textbf{0.001} & \textbf{0.043} & 0.191 & \textbf{0.001} & \textbf{0.001} & \textbf{0.001} & \textbf{0.001} & \textbf{0.001} & \textbf{0.001} & 0.170 & - & \textbf{0.002} \\
        Weighting $\lambda$=10 & \textbf{0.002} & \textbf{0.001} & \textbf{0.001} & \textbf{0.001} & 0.068 & \textbf{0.001} & \textbf{0.001} & \textbf{0.001} & \textbf{0.003} & \textbf{0.001} & \textbf{0.001} & \textbf{0.001} & \textbf{0.002} & - \\
        \hline
    \end{tabular}
    \label{tab:fine-tuning-signifficance-comet}
\end{table*}

\begin{table*}[h]
    \centering
    \scriptsize
    \caption{The p-values of the paired bootstrapping of the results in terms of ctxPro accuracy of the Gender phenomenon for each pair of the models based on OpusMT en-de. Values <0.05 are in bold.}
    \begin{tabular}{l|cccccccccccccc}
        \hline
        \textbf{Model} & \rotatebox[origin=c]{90}{Baseline} & \rotatebox[origin=c]{90}{CoWord p=0.1} & \rotatebox[origin=c]{90}{CoWord p=0.2} & \rotatebox[origin=c]{90}{CoWord p=0.3} & \rotatebox[origin=c]{90}{Adapted D\&R} & \rotatebox[origin=c]{90}{Fine-tuning e=1} & \rotatebox[origin=c]{90}{Fine-tuning e=2} & \rotatebox[origin=c]{90}{Fine-tuning e=5} & \rotatebox[origin=c]{90}{MaxPCXMI e=1} & \rotatebox[origin=c]{90}{MaxPCXMI e=2} & \rotatebox[origin=c]{90}{ MaxPCXMI e=5 } & \rotatebox[origin=c]{90}{Weighting $\lambda$=2} & \rotatebox[origin=c]{90}{Weighting $\lambda$=5} & \rotatebox[origin=c]{90}{Weighting $\lambda$=10} \\
        \hline
        Baseline & - & 0.159 & \textbf{0.014} & \textbf{0.001} & \textbf{0.001} & \textbf{0.001} & \textbf{0.001} & \textbf{0.001} & \textbf{0.001} & \textbf{0.001} & \textbf{0.001} & \textbf{0.001} & \textbf{0.001} & \textbf{0.001} \\
        CoWord p=0.1 & 0.159 & - & \textbf{0.011} & \textbf{0.001} & \textbf{0.001} & \textbf{0.001} & \textbf{0.001} & \textbf{0.001} & \textbf{0.001} & \textbf{0.001} & \textbf{0.001} & \textbf{0.001} & \textbf{0.001} & \textbf{0.001} \\
        CoWord p=0.2 & \textbf{0.014} & \textbf{0.011} & - & \textbf{0.001} & \textbf{0.001} & \textbf{0.001} & \textbf{0.001} & \textbf{0.001} & \textbf{0.001} & \textbf{0.001} & \textbf{0.001} & \textbf{0.001} & \textbf{0.001} & \textbf{0.001} \\
        CoWord p=0.3 & \textbf{0.001} & \textbf{0.001} & \textbf{0.001} & - & \textbf{0.001} & \textbf{0.001} & \textbf{0.001} & \textbf{0.001} & \textbf{0.001} & \textbf{0.001} & \textbf{0.001} & \textbf{0.001} & \textbf{0.001} & \textbf{0.001} \\
        Adapted D\&R & \textbf{0.001} & \textbf{0.001} & \textbf{0.001} & \textbf{0.001} & - & \textbf{0.001} & \textbf{0.001} & \textbf{0.001} & \textbf{0.001} & \textbf{0.001} & \textbf{0.001} & \textbf{0.001} & \textbf{0.001} & \textbf{0.001} \\
        Fine-tuning e=1 & \textbf{0.001} & \textbf{0.001} & \textbf{0.001} & \textbf{0.001} & \textbf{0.001} & - & \textbf{0.001} & \textbf{0.001} & \textbf{0.001} & 0.058 & \textbf{0.001} & \textbf{0.001} & \textbf{0.001} & \textbf{0.001} \\
        Fine-tuning e=2 & \textbf{0.001} & \textbf{0.001} & \textbf{0.001} & \textbf{0.001} & \textbf{0.001} & \textbf{0.001} & - & \textbf{0.001} & \textbf{0.001} & \textbf{0.001} & \textbf{0.002} & \textbf{0.001} & \textbf{0.001} & \textbf{0.001} \\
        Fine-tuning e=5 & \textbf{0.001} & \textbf{0.001} & \textbf{0.001} & \textbf{0.001} & \textbf{0.001} & \textbf{0.001} & \textbf{0.001} & - & \textbf{0.001} & \textbf{0.001} & \textbf{0.001} & \textbf{0.001} & \textbf{0.001} & \textbf{0.001} \\
        MaxPCXMI e=1 & \textbf{0.001} & \textbf{0.001} & \textbf{0.001} & \textbf{0.001} & \textbf{0.001} & \textbf{0.001} & \textbf{0.001} & \textbf{0.001} & - & \textbf{0.001} & \textbf{0.001} & \textbf{0.001} & \textbf{0.001} & 0.166 \\
        MaxPCXMI e=2 & \textbf{0.001} & \textbf{0.001} & \textbf{0.001} & \textbf{0.001} & \textbf{0.001} & 0.058 & \textbf{0.001} & \textbf{0.001} & \textbf{0.001} & - & \textbf{0.001} & \textbf{0.001} & \textbf{0.001} & \textbf{0.001} \\
        MaxPCXMI e=5 & \textbf{0.001} & \textbf{0.001} & \textbf{0.001} & \textbf{0.001} & \textbf{0.001} & \textbf{0.001} & \textbf{0.002} & \textbf{0.001} & \textbf{0.001} & \textbf{0.001} & - & \textbf{0.001} & \textbf{0.001} & \textbf{0.001} \\
        Weighting $\lambda$=2 & \textbf{0.001} & \textbf{0.001} & \textbf{0.001} & \textbf{0.001} & \textbf{0.001} & \textbf{0.001} & \textbf{0.001} & \textbf{0.001} & \textbf{0.001} & \textbf{0.001} & \textbf{0.001} & - & \textbf{0.001} & \textbf{0.001} \\
        Weighting $\lambda$=5 & \textbf{0.001} & \textbf{0.001} & \textbf{0.001} & \textbf{0.001} & \textbf{0.001} & \textbf{0.001} & \textbf{0.001} & \textbf{0.001} & \textbf{0.001} & \textbf{0.001} & \textbf{0.001} & \textbf{0.001} & - & \textbf{0.001} \\
        Weighting $\lambda$=10 & \textbf{0.001} & \textbf{0.001} & \textbf{0.001} & \textbf{0.001} & \textbf{0.001} & \textbf{0.001} & \textbf{0.001} & \textbf{0.001} & 0.166 & \textbf{0.001} & \textbf{0.001} & \textbf{0.001} & \textbf{0.001} & - \\
        \hline
    \end{tabular}
    \label{tab:fine-tuning-signifficance-gender}
\end{table*}

\begin{table*}[h]
    \centering
    \scriptsize
    \caption{The p-values of the paired bootstrapping of the results in terms of ctxPro accuracy of the Formality phenomenon for each pair of the models based on OpusMT en-de. Values <0.05 are in bold.}
    \begin{tabular}{l|cccccccccccccc}
        \hline
        \textbf{Model} & \rotatebox[origin=c]{90}{Baseline} & \rotatebox[origin=c]{90}{CoWord p=0.1} & \rotatebox[origin=c]{90}{CoWord p=0.2} & \rotatebox[origin=c]{90}{CoWord p=0.3} & \rotatebox[origin=c]{90}{Adapted D\&R} & \rotatebox[origin=c]{90}{Fine-tuning e=1} & \rotatebox[origin=c]{90}{Fine-tuning e=2} & \rotatebox[origin=c]{90}{Fine-tuning e=5} & \rotatebox[origin=c]{90}{MaxPCXMI e=1} & \rotatebox[origin=c]{90}{MaxPCXMI e=2} & \rotatebox[origin=c]{90}{ MaxPCXMI e=5 } & \rotatebox[origin=c]{90}{Weighting $\lambda$=2} & \rotatebox[origin=c]{90}{Weighting $\lambda$=5} & \rotatebox[origin=c]{90}{Weighting $\lambda$=10} \\
        \hline
        Baseline & - & \textbf{0.002} & \textbf{0.002} & \textbf{0.027} & \textbf{0.001} & \textbf{0.001} & \textbf{0.001} & \textbf{0.001} & \textbf{0.001} & \textbf{0.001} & \textbf{0.001} & \textbf{0.001} & \textbf{0.001} & \textbf{0.001} \\
        CoWord p=0.1 & \textbf{0.002} & - & 0.269 & 0.110 & \textbf{0.001} & \textbf{0.001} & \textbf{0.001} & \textbf{0.001} & \textbf{0.001} & \textbf{0.001} & \textbf{0.001} & \textbf{0.001} & \textbf{0.001} & \textbf{0.001} \\
        CoWord p=0.2 & \textbf{0.002} & 0.269 & - & \textbf{0.019} & \textbf{0.001} & \textbf{0.001} & \textbf{0.001} & \textbf{0.001} & \textbf{0.001} & \textbf{0.001} & \textbf{0.001} & \textbf{0.001} & \textbf{0.001} & \textbf{0.001} \\
        CoWord p=0.3 & \textbf{0.027} & 0.110 & \textbf{0.019} & - & \textbf{0.001} & \textbf{0.001} & \textbf{0.001} & \textbf{0.001} & \textbf{0.001} & \textbf{0.001} & \textbf{0.001} & \textbf{0.001} & \textbf{0.001} & \textbf{0.001} \\
        Adapted D\&R & \textbf{0.001} & \textbf{0.001} & \textbf{0.001} & \textbf{0.001} & - & \textbf{0.001} & \textbf{0.001} & \textbf{0.001} & \textbf{0.001} & \textbf{0.001} & \textbf{0.001} & \textbf{0.001} & \textbf{0.001} & \textbf{0.001} \\
        Fine-tuning e=1 & \textbf{0.001} & \textbf{0.001} & \textbf{0.001} & \textbf{0.001} & \textbf{0.001} & - & 0.057 & \textbf{0.001} & \textbf{0.001} & \textbf{0.001} & \textbf{0.001} & \textbf{0.031} & \textbf{0.017} & \textbf{0.001} \\
        Fine-tuning e=2 & \textbf{0.001} & \textbf{0.001} & \textbf{0.001} & \textbf{0.001} & \textbf{0.001} & 0.057 & - & \textbf{0.001} & \textbf{0.001} & \textbf{0.001} & \textbf{0.001} & \textbf{0.015} & \textbf{0.045} & \textbf{0.001} \\
        Fine-tuning e=5 & \textbf{0.001} & \textbf{0.001} & \textbf{0.001} & \textbf{0.001} & \textbf{0.001} & \textbf{0.001} & \textbf{0.001} & - & \textbf{0.001} & \textbf{0.001} & \textbf{0.001} & \textbf{0.001} & 0.356 & \textbf{0.001} \\
        MaxPCXMI e=1 & \textbf{0.001} & \textbf{0.001} & \textbf{0.001} & \textbf{0.001} & \textbf{0.001} & \textbf{0.001} & \textbf{0.001} & \textbf{0.001} & - & \textbf{0.001} & \textbf{0.001} & \textbf{0.001} & \textbf{0.001} & \textbf{0.001} \\
        MaxPCXMI e=2 & \textbf{0.001} & \textbf{0.001} & \textbf{0.001} & \textbf{0.001} & \textbf{0.001} & \textbf{0.001} & \textbf{0.001} & \textbf{0.001} & \textbf{0.001} & - & \textbf{0.001} & \textbf{0.001} & \textbf{0.001} & \textbf{0.001} \\
        MaxPCXMI e=5 & \textbf{0.001} & \textbf{0.001} & \textbf{0.001} & \textbf{0.001} & \textbf{0.001} & \textbf{0.001} & \textbf{0.001} & \textbf{0.001} & \textbf{0.001} & \textbf{0.001} & - & \textbf{0.001} & \textbf{0.001} & \textbf{0.001} \\
        Weighting $\lambda$=2 & \textbf{0.001} & \textbf{0.001} & \textbf{0.001} & \textbf{0.001} & \textbf{0.001} & \textbf{0.031} & \textbf{0.015} & \textbf{0.001} & \textbf{0.001} & \textbf{0.001} & \textbf{0.001} & - & \textbf{0.001} & \textbf{0.001} \\
        Weighting $\lambda$=5 & \textbf{0.001} & \textbf{0.001} & \textbf{0.001} & \textbf{0.001} & \textbf{0.001} & \textbf{0.017} & \textbf{0.045} & 0.356 & \textbf{0.001} & \textbf{0.001} & \textbf{0.001} & \textbf{0.001} & - & \textbf{0.001} \\
        Weighting $\lambda$=10 & \textbf{0.001} & \textbf{0.001} & \textbf{0.001} & \textbf{0.001} & \textbf{0.001} & \textbf{0.001} & \textbf{0.001} & \textbf{0.001} & \textbf{0.001} & \textbf{0.001} & \textbf{0.001} & \textbf{0.001} & \textbf{0.001} & - \\
        \hline
    \end{tabular}
    \label{tab:fine-tuning-signifficance-formality}
\end{table*}

\begin{table*}[h]
    \centering
    \scriptsize
    \caption{The p-values of the paired bootstrapping of the results in terms of ctxPro accuracy of the Auxiliary phenomenon for each pair of the models based on OpusMT en-de. Values <0.05 are in bold.}
    \begin{tabular}{l|cccccccccccccc}
        \hline
        \textbf{Model} & \rotatebox[origin=c]{90}{Baseline} & \rotatebox[origin=c]{90}{CoWord p=0.1} & \rotatebox[origin=c]{90}{CoWord p=0.2} & \rotatebox[origin=c]{90}{CoWord p=0.3} & \rotatebox[origin=c]{90}{Adapted D\&R} & \rotatebox[origin=c]{90}{Fine-tuning e=1} & \rotatebox[origin=c]{90}{Fine-tuning e=2} & \rotatebox[origin=c]{90}{Fine-tuning e=5} & \rotatebox[origin=c]{90}{MaxPCXMI e=1} & \rotatebox[origin=c]{90}{MaxPCXMI e=2} & \rotatebox[origin=c]{90}{ MaxPCXMI e=5 } & \rotatebox[origin=c]{90}{Weighting $\lambda$=2} & \rotatebox[origin=c]{90}{Weighting $\lambda$=5} & \rotatebox[origin=c]{90}{Weighting $\lambda$=10} \\
        \hline
        Baseline & - & \textbf{0.001} & \textbf{0.001} & \textbf{0.001} & \textbf{0.031} & \textbf{0.001} & \textbf{0.001} & \textbf{0.010} & 0.364 & 0.108 & \textbf{0.012} & \textbf{0.002} & \textbf{0.001} & \textbf{0.001} \\
        CoWord p=0.1 & \textbf{0.001} & - & \textbf{0.001} & \textbf{0.001} & \textbf{0.001} & \textbf{0.001} & \textbf{0.001} & \textbf{0.001} & \textbf{0.001} & \textbf{0.001} & \textbf{0.001} & \textbf{0.001} & 0.135 & \textbf{0.016} \\
        CoWord p=0.2 & \textbf{0.001} & \textbf{0.001} & - & \textbf{0.001} & \textbf{0.001} & \textbf{0.001} & \textbf{0.001} & \textbf{0.001} & \textbf{0.001} & \textbf{0.001} & \textbf{0.001} & \textbf{0.001} & \textbf{0.001} & 0.155 \\
        CoWord p=0.3 & \textbf{0.001} & \textbf{0.001} & \textbf{0.001} & - & \textbf{0.001} & \textbf{0.001} & \textbf{0.001} & \textbf{0.001} & \textbf{0.001} & \textbf{0.001} & \textbf{0.001} & \textbf{0.001} & \textbf{0.001} & \textbf{0.002} \\
        Adapted D\&R & \textbf{0.031} & \textbf{0.001} & \textbf{0.001} & \textbf{0.001} & - & \textbf{0.001} & \textbf{0.001} & \textbf{0.001} & 0.103 & 0.352 & 0.109 & \textbf{0.046} & \textbf{0.001} & \textbf{0.001} \\
        Fine-tuning e=1 & \textbf{0.001} & \textbf{0.001} & \textbf{0.001} & \textbf{0.001} & \textbf{0.001} & - & 0.097 & \textbf{0.006} & \textbf{0.001} & \textbf{0.001} & \textbf{0.001} & \textbf{0.001} & \textbf{0.001} & \textbf{0.001} \\
        Fine-tuning e=2 & \textbf{0.001} & \textbf{0.001} & \textbf{0.001} & \textbf{0.001} & \textbf{0.001} & 0.097 & - & \textbf{0.029} & \textbf{0.001} & \textbf{0.001} & \textbf{0.001} & \textbf{0.001} & \textbf{0.001} & \textbf{0.001} \\
        Fine-tuning e=5 & \textbf{0.010} & \textbf{0.001} & \textbf{0.001} & \textbf{0.001} & \textbf{0.001} & \textbf{0.006} & \textbf{0.029} & - & \textbf{0.001} & \textbf{0.001} & \textbf{0.001} & \textbf{0.001} & \textbf{0.001} & \textbf{0.001} \\
        MaxPCXMI e=1 & 0.364 & \textbf{0.001} & \textbf{0.001} & \textbf{0.001} & 0.103 & \textbf{0.001} & \textbf{0.001} & \textbf{0.001} & - & \textbf{0.005} & \textbf{0.001} & \textbf{0.017} & \textbf{0.001} & \textbf{0.001} \\
        MaxPCXMI e=2 & 0.108 & \textbf{0.001} & \textbf{0.001} & \textbf{0.001} & 0.352 & \textbf{0.001} & \textbf{0.001} & \textbf{0.001} & \textbf{0.005} & - & \textbf{0.002} & 0.092 & \textbf{0.001} & \textbf{0.001} \\
        MaxPCXMI e=5 & \textbf{0.012} & \textbf{0.001} & \textbf{0.001} & \textbf{0.001} & 0.109 & \textbf{0.001} & \textbf{0.001} & \textbf{0.001} & \textbf{0.001} & \textbf{0.002} & - & 0.421 & \textbf{0.003} & \textbf{0.001} \\
        Weighting $\lambda$=2 & \textbf{0.002} & \textbf{0.001} & \textbf{0.001} & \textbf{0.001} & \textbf{0.046} & \textbf{0.001} & \textbf{0.001} & \textbf{0.001} & \textbf{0.017} & 0.092 & 0.421 & - & \textbf{0.001} & \textbf{0.001} \\
        Weighting $\lambda$=5 & \textbf{0.001} & 0.135 & \textbf{0.001} & \textbf{0.001} & \textbf{0.001} & \textbf{0.001} & \textbf{0.001} & \textbf{0.001} & \textbf{0.001} & \textbf{0.001} & \textbf{0.003} & \textbf{0.001} & - & \textbf{0.001} \\
        Weighting $\lambda$=10 & \textbf{0.001} & \textbf{0.016} & 0.155 & \textbf{0.002} & \textbf{0.001} & \textbf{0.001} & \textbf{0.001} & \textbf{0.001} & \textbf{0.001} & \textbf{0.001} & \textbf{0.001} & \textbf{0.001} & \textbf{0.001} & - \\
        \hline
    \end{tabular}
    \label{tab:fine-tuning-signifficance-auxiliary}
\end{table*}

\end{document}